\documentclass[10pt,twocolumn,letterpaper]{article}

\usepackage{wacv}
\usepackage{graphicx}

\usepackage{bm}
\usepackage{tabularx}
\usepackage{booktabs}
\usepackage{multirow}
\usepackage{tikz}
\usepackage{pgfplots}
\usepackage{twemojis}
\usepackage{amssymb}
\usepackage{makecell}

\newcolumntype{Y}{>{\centering\arraybackslash}X}

\definecolor{c_red}{HTML}{ea4335}
\definecolor{c_green}{HTML}{34a853}
\definecolor{c_idil}{HTML}{8e44ad}
\definecolor{c_bastian}{HTML}{2e86de}

\colorlet{tablered}{c_red!99!black}
\colorlet{tablegreen}{c_green!99!black}

\newcommand{\cmark}{\checkmark}
\newcommand{\xmark}{$\times$}

\newcommand{\PAR}[1]{\vskip4pt \noindent {\bf #1~}}
\newcommand{\PARbegin}[1]{\vskip1pt \noindent {\bf #1~}}

\newcommand{\tablestep}[1]{\textcolor{gray!90}{(#1)}}

\newcommand{\oursShort}{LVMT}
\newcommand{\ours}{Long-term Video Mask Transformer}
\newcommand{\tqp}{Truncated Query Propagation}
\newcommand{\tqpshort}{TQP}

\makeatletter
\newcommand\notsotiny{\@setfontsize\notsotiny{6.5}{7.5}}
\newcommand\notscriptsize{\@setfontsize\notscriptsize{7.5}{8.5}}
\makeatother

\usepackage{microtype}

\usepackage{pgfplots}
\pgfplotsset{compat=1.18}
\usetikzlibrary{arrows.meta, backgrounds, calc}
\usepackage{fontawesome5}

\expandafter\def\expandafter\normalsize\expandafter{\normalsize\setlength\abovedisplayskip{2pt}\setlength\belowdisplayskip{2pt}\setlength\abovedisplayshortskip{2pt}\setlength\belowdisplayshortskip{2pt}}

\renewcommand{\arraystretch}{0.9}
\usepackage{xcolor}

\definecolor{wacvblue}{rgb}{0.21,0.49,0.74}
\usepackage[pagebackref,breaklinks,colorlinks,allcolors=wacvblue]{hyperref}

\def\wacvPaperID{93}
\def\confName{WACV}
\def\confYear{2027}

\title{LVMT: Video Mask Transformer for Long-term Video Segmentation}

\author{
Narges Norouzi\textsuperscript{1}\qquad 
Niccolò Cavagnero\textsuperscript{1} \qquad 
Idil Esen Zulfikar\textsuperscript{2} \qquad 
Bastian Leibe\textsuperscript{2} \qquad \\
Gijs Dubbelman\textsuperscript{1} \qquad 
Daan de Geus\textsuperscript{1}\\[0.6em]
$^1$Eindhoven University of Technology\qquad $^2$RWTH Aachen University \\
}

\begin{document}

\maketitle
\begin{abstract}
\label{sec:abstract}
Existing online video segmentation methods struggle to track objects in long, complex videos with long-term occlusions.
We hypothesize that this limitation is caused by (i) the inability of their temporal propagation mechanism to adaptively select the object information that is propagated across time, and (ii) their inability to be trained on long videos due to memory requirements and vanishing gradients.
To address the first limitation, we propose to use a lightweight GRU-based temporal propagation module that can learn to select which information it keeps in memory and propagates across time.
Second, to allow training on long videos, we leverage \tqp{}~(\tqpshort{}), a training strategy in which the model processes a video in chunks of frames, where information about tracked objects is propagated between chunks but backpropagation is only conducted in individual chunks, enabling longer temporal supervision without out-of-memory issues, inference overhead, or vanishing gradients. 
The resulting model is called the \ours{} (\oursShort{}).
Extensive experiments on six benchmarks show that \oursShort{} sets a new state of the art across a range of video segmentation tasks, while retaining the speed of the highly efficient model it is based on, making it 10$\times$ faster than the prior state of the art.
Code: \href{https://github.com/tue-mps/lvmt}{https://github.com/tue-mps/lvmt}.
\vspace{-5pt}

\end{abstract}

\section{Introduction}
\label{sec:intro}

\definecolor{pmtcolor}{HTML}{397FBE}
\definecolor{lvmtcolor}{HTML}{FF6D01}

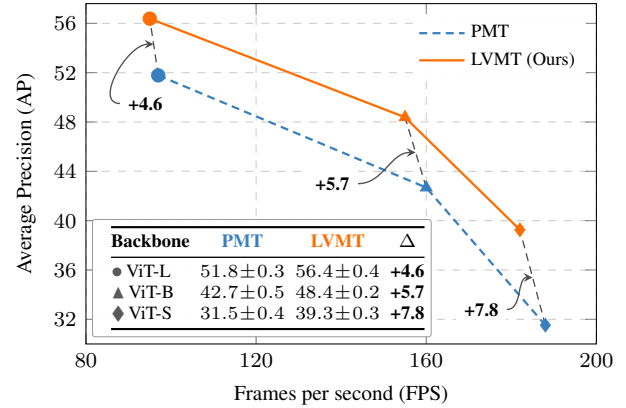
\begin{figure}[t!]
\centering
\begin{tikzpicture}
\begin{axis}[
width=1.0\linewidth,
height=0.73\linewidth,
font=\small,
xlabel={Frames per second (FPS)},
ylabel={Average Precision (AP)},
xmin=80,
xmax=200,
xtick={80,120,160,200},
ymin=30,
ymax=57.6,
ytick={32,36,40,44,48,52,56},
xmajorgrids=true,
ymajorgrids=true,
grid style={dashed,black!20},
tick label style={font=\footnotesize},
label style={font=\footnotesize},
legend entries={PMT,\oursShort{} (Ours)},
legend style={font=\scriptsize,at={(0.98,0.97)},anchor=north east,draw=none,fill=white,fill opacity=0.95,text opacity=1,inner sep=2pt},
legend cell align=left,
clip=false
]
\addplot[color=pmtcolor,densely dashed,line width=0.3mm,mark=none] coordinates {(97,51.78)(160,42.68)(188,31.52)};
\addplot[color=pmtcolor,only marks,mark=*,mark size=2.5pt,forget plot] coordinates {(97,51.78)};
\addplot[color=pmtcolor,only marks,mark=triangle*,mark size=2.5pt,forget plot] coordinates {(160,42.68)};
\addplot[color=pmtcolor,only marks,mark=diamond*,mark size=2.5pt,forget plot] coordinates {(188,31.52)};
\addplot[color=lvmtcolor,line width=0.3mm,mark=none] coordinates {(95,56.38)(155,48.38)(182,39.26)};
\addplot[color=lvmtcolor,only marks,mark=*,mark size=2.5pt,forget plot] coordinates {(95,56.38)};
\addplot[color=lvmtcolor,only marks,mark=triangle*,mark size=2.5pt,forget plot] coordinates {(155,48.38)};
\addplot[color=lvmtcolor,only marks,mark=diamond*,mark size=2.5pt,forget plot] coordinates {(182,39.26)};

\draw[densely dashed,black!70,line width=0.5pt] (axis cs:97,51.78) -- coordinate[pos=0.55] (midL) (axis cs:95,56.38);
\node[font=\scriptsize\bfseries,fill=white,inner sep=3pt,anchor=west] (labL) at (axis cs:88,49.5) {+4.6};
\draw[-stealth,black!70,line width=0.5pt] (labL.west) to[out=180,in=180] (midL);

\draw[densely dashed,black!70,line width=0.5pt] (axis cs:160,42.68) -- coordinate[pos=0.5] (midB) (axis cs:155,48.38);
\node[font=\scriptsize\bfseries,fill=white,inner sep=3pt,anchor=west] (labB) at (axis cs:132,43.0) {+5.7};
\draw[-stealth,black!70,line width=0.5pt] (labB.east) to[out=0,in=180] (midB);

\draw[densely dashed,black!70,line width=0.5pt] (axis cs:188,31.52) -- coordinate[pos=0.5] (midS) (axis cs:182,39.26);
\node[font=\scriptsize\bfseries,fill=white,inner sep=3pt,anchor=west] (labS) at (axis cs:167,32.8) {+7.8};
\draw[-stealth,black!70,line width=0.5pt] (labS.east) to[out=0,in=180] (midS);

\node[
    anchor=south west,
    fill=white,
    fill opacity=0.95,
    text opacity=1,
    draw=black!40,
    line width=0.3pt,
    rounded corners=1pt,
    inner sep=2.5pt
] at (rel axis cs:0.012,0.02) {
    \renewcommand{\arraystretch}{0.98}
    \setlength{\tabcolsep}{2pt}
    \scriptsize
    \begin{tabular}{@{}lccc@{}}
        \toprule
        \textbf{Backbone} &
        \textcolor{pmtcolor}{\textbf{PMT}} &
        \textcolor{lvmtcolor}{\textbf{\oursShort{}}} &
        \textbf{$\Delta$} \\
        \midrule
        \textcolor{black!65}{\scalebox{1.1}{$\bullet$}}\,ViT-L &
        $51.8\!\pm\!0.3$ &
        $56.4\!\pm\!0.4$ &
        \textbf{+4.6} \\
        \textcolor{black!65}{$\blacktriangle$}\,ViT-B &
        $42.7\!\pm\!0.5$ &
        $48.4\!\pm\!0.2$ &
        \textbf{+5.7} \\
        \textcolor{black!65}{$\blacklozenge$}\,ViT-S &
        $31.5\!\pm\!0.4$ &
        $39.3\!\pm\!0.3$ &
        \textbf{+7.8} \\
        \bottomrule
    \end{tabular}
};

\end{axis}
\end{tikzpicture}
\caption{
\textbf{PMT vs. \oursShort{}.}
Mean AP $\pm$ std.~dev.~over five runs. Across ViT-L/B/S,
\oursShort{} improves AP by at least $\mathbf{+4.6}$ over the efficient PMT~\cite{cavagnero2026pmt} baseline at similar FPS. Evaluated on OVIS \textit{val}.
}
\vspace{-1pt}
\label{fig:teaser}
\end{figure}

Video segmentation refers to the task of segmenting, classifying, and tracking object instances consistently across all frames of a video sequence.
Recent video segmentation approaches VidEoMT~\cite{norouzi2026videomt} and PMT~\cite{cavagnero2026pmt} show that large, extensively pre-trained Vision Transformers (ViT) encoders can replace the many complex components that are commonly used in earlier models~\cite{lee2025cavis,zhou2024dvisdaq,zhang2025dvis++}, resulting in much simpler and faster architectures that obtain competitive, state-of-the-art accuracy.
VidEoMT employs an encoder-only segmentation model~\cite{kerssies2025eomt} for each video frame, and achieves tracking by propagating \emph{query embeddings} with information about the previous frame's objects to the model for the next time step. 
To allow the pre-trained encoder to be reused and support multiple tasks in parallel, PMT instead leverages a lightweight decoder that is applied on top of a frozen ViT, with temporal modeling working the same as for VidEoMT.
Despite their high accuracy and efficiency, these models still struggle with long-term object tracking, just like prior methods.
Especially in long, complex videos and under long-term occlusion, their predictions exhibit erroneous \emph{identity switches}, \ie, inconsistent object identity assignment across frames.
The objective of this paper is to improve the accuracy of these models by better modeling long-range temporal information while preserving the simplicity and speed of current efficient architectures.

A key limitation of these existing efficient models is that their temporal query propagation mechanism cannot adaptively select which information about objects it propagates across time.
The propagated queries are always the sum of temporally-agnostic learnable queries and per-object query embeddings from the previous frame.
In case of long occlusions, information about occluded objects is eventually diluted by the iterative addition of the temporally-agnostic queries, causing the model to struggle to re-identify these occluded objects.
A straightforward solution would be to use an explicit external memory bank with the queries for all tracked objects~\cite{heo2023generalized,wu2022defense}, including the occluded ones.
However, such a growing query history increases both memory usage and computation with video length, resulting in poor model efficiency when applied to long videos.

Instead of a growing query history, we propose to use a lightweight learnable memory implemented as a GRU cell~\cite{cho2014gru}. 
In this recurrent module, the model can adaptively select which information from the previous time step it keeps in memory and propagates to the next time step.
With this, we hypothesize that the model can learn to keep information about occluded objects in memory when it needs to, allowing for re-identification when these objects reappear.

Although the recurrent unit improves overall performance, we empirically observe that identity switches remain a key failure mode in challenging videos. 
From our analysis, reported in \cref{sec:from_pmt_to_vmt}, we find that the model suffers from identity switches on videos that contain many objects, frequent disappearances, and long-term occlusions. Since the model is trained only on short clips, which is the default for PMT~\cite{cavagnero2026pmt}, such gaps exceed its training horizon, complicating identity recovery at test time. 
A natural solution would be to train on longer video clips. 
However, this introduces two practical challenges. 
First, processing more frames substantially increases memory consumption and can lead to out-of-memory errors.
Second, optimizing recurrent updates over long horizons causes vanishing gradients through time, yielding worse performance.

To address these limitations, we adopt Truncated Query Propagation (\tqpshort{}), a training strategy inspired by TBPTT~\cite{williams1990tbptt}. 
With \tqpshort{}, each video is divided into chunks, and queries are propagated forward across chunk boundaries to preserve temporal context. 
During backpropagation, gradients are truncated within each chunk, so gradients from temporally distant predictions are not propagated through the full sequence.
This allows the model to learn longer-term temporal dependencies during training without suffering from vanishing gradients or memory issues.

Applying the GRU-based memory and \tqpshort{} to PMT, we present the \textit{\ours{}} (\oursShort{}) model, which better incorporates and preserves long-term temporal information to improve accuracy without compromising efficiency.~\oursShort{} offers several advantages: $(i)$ the fixed-size GRU state enables long-range temporal modeling without growing memory usage or computation; $(ii)$ chunk-based training with \tqpshort{} allows training on arbitrarily long videos without memory issues; $(iii)$ \tqpshort{} stabilizes gradient propagation through the recurrent GRU updates by truncating the backward pass over time; $(iv)$ as a training-time strategy, \tqpshort{} improves accuracy without adding inference-time overhead and without altering the model architecture.

Our comprehensive experimental analysis demonstrates that \oursShort{} consistently outperforms PMT while maintaining nearly identical efficiency across all benchmarks.
Notably, on OVIS~\cite{qi2022occluded}, a benchmark specifically designed to stress-test models under heavy, long-term occlusion, \oursShort{} improves AP by at least $+4.6$ over PMT across ViT-L/B/S backbones at similar prediction speeds (see Fig.~\ref{fig:teaser}).
Moreover, \oursShort{} surpasses the prior state-of-the-art method DVIS-DAQ~\cite{zhou2024dvisdaq} by $+2.4$ AP while retaining PMT's efficiency, enabling it to be over $10\times$ faster than DVIS-DAQ.
These gains are further validated on other benchmarks for video instance, panoptic, and semantic segmentation in ~\cref{sec:experiments}.

In summary, we make the following contributions:
\begin{itemize}
\item We introduce a lightweight GRU-based temporal propagation mechanism for video segmentation that captures long-range context while avoiding both growing memory usage over time and excessive computational overhead.
\item 
We leverage Truncated Query Propagation (TQP), which adapts TBPTT with across-chunk gradient accumulation to query propagation, enabling long-video training with bounded peak memory and improved gradient stability.

\item The resulting model, \oursShort{}, achieves a stronger performance \vs latency trade-off than prior methods.
\end{itemize}

\section{Related Work}
\label{sec:related_work}

\PAR{Video Segmentation.} Current state-of-the-art video segmentation methods~\cite{huang2022minvis, zhang2023dvis, zhang2025dvis++,zhou2024dvisdaq,lee2025cavis, lee2025lomm, ke2022video} are universal models, which means that they use a single framework for video instance segmentation (VIS)~\cite{yang2019video}, video panoptic segmentation (VPS)~\cite{kim2020video}, \emph{and} video semantic segmentation (VSS)~\cite{nilsson2018semantic}. 
These methods typically follow a decoupled paradigm, using a segmenter for per-frame segmentation and a tracker for temporal association. While effective, these specialized components increase the complexity of the architecture and severely reduce their efficiency. 

To improve efficiency without harming accuracy, recent work introduces VidEoMT~\cite{norouzi2026videomt}, an encoder-only video segmentation method that replaces complex components with a large, extensively pre-trained ViT encoder and simple temporal query propagation.
The follow-up method PMT~\cite{cavagnero2026pmt} extends VidEoMT to the setting where the ViT encoder remains frozen and can be reused for various downstream tasks, by introducing a lightweight decoder. While these models are effective, they still struggle with long-term tracking due to the non-adaptive query propagation mechanism and their inability to train effectively on long videos.
This work addresses both limitations.

\PAR{Long-term Temporal Modeling.} 
Several works explore long-term temporal modeling for object tracking and video segmentation. In multiple object tracking, TrackFormer~\cite{meinhardt2022trackformer} keeps a short-term history of tracked objects and aims to re-identify those in future frames. However, Norouzi \etal~\cite{norouzi2026videomt} show that such propagation is inefficient for video segmentation, as it requires applying non-maximum suppression over masks to remove duplicate queries at each frame.
GenVIS~\cite{heo2023generalized} introduces an explicit query memory for VIS. Unlike TrackFormer, it retains object queries indefinitely, enabling long-term association. However, this comes at the cost of increasing memory and computation as video length grows, due to repeated cross-attention over stored queries. Recently, several methods, such as  LiVOS~\cite{liu2025livos}, XMem~\cite{cheng2022xmem}, and Cutie~\cite{cheng2024putting}, explored long-term temporal modeling for video \emph{object} segmentation (VOS). Designed for VOS, they rely on multiple explicit memory banks and iterative memory retrieval, resulting in computational costs that scale poorly with video length and object count and making them unsuitable for efficient application to VIS, VPS, and VSS.

More recently, SAM3~\cite{carion2025sam3segmentconcepts} achieves strong performance in promptable video segmentation and tracking. However, it relies on a complex architecture comprising vision and text encoders together with multiple task-specific modules for detection, segmentation, and tracking, many of which are similar to the ones shown to be highly inefficient by Norouzi \etal~\cite{norouzi2026videomt}. Moreover, as the number of tracked entities increases, the model speed consistently decreases.

In contrast, we aim for \emph{efficient} long-term modeling and propose a lightweight GRU-based propagation mechanism with truncated chunk-based training, enabling long-horizon consistency at the high efficiency of VidEoMT and PMT.

\begin{figure*}[t]
\centering
\includegraphics[width=0.98\linewidth]{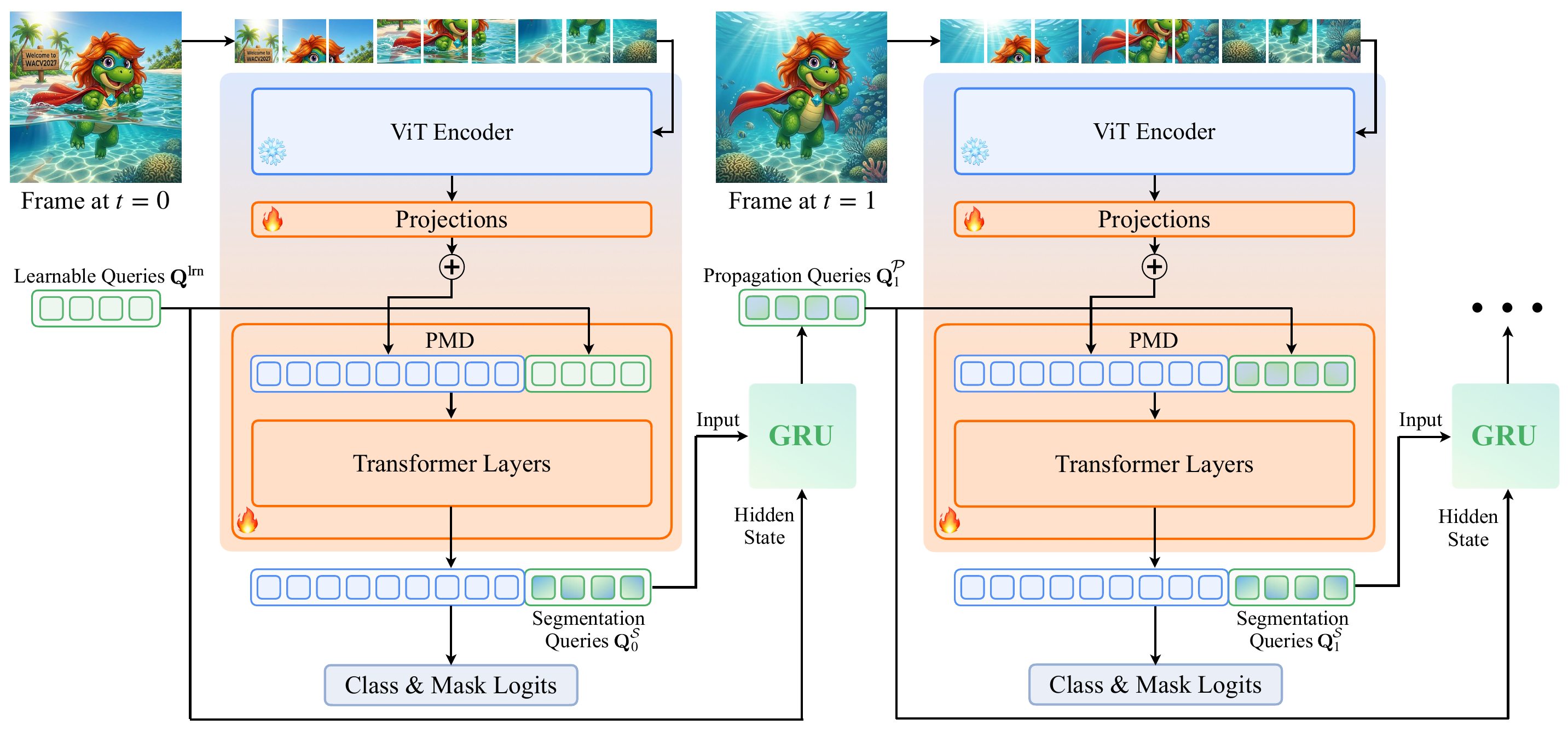}
\caption{\textbf{\oursShort{} architecture.} The Plain Mask Decoder (PMD) takes input queries and projected patch features from a frozen ViT encoder to produce segmentation queries for mask and class prediction. At $t=0$, learnable queries $\mathbf{Q}^{\mathrm{lrn}}$ are used to initialize both the decoder input and the GRU hidden state. Thereafter, a GRU cell adaptively updates the hidden state from the current segmentation queries to yield propagation queries for the next frame. Truncated Query Propagation (TQP), visualized in~\cref{fig:tqp}, is applied at training time.}
\label{fig:arch}
\end{figure*}

\section{Preliminaries}
\label{sec:prelim}

\PARbegin{Task Definition.}
This work focuses on online video segmentation: 
at each time step, the model produces a segmentation mask and category label for each object in the frame, and it re-identifies objects that were detected in previous frames. In this work, we use ``object'' as a general term that may refer to object instances in VIS, semantic classes in VSS, or both in VPS.
Formally, given a video $\mathcal{V} = \{\mathbf{I}_1, \mathbf{I}_2, \dots, \mathbf{I}_T\}$ of $T$ frames, for each frame $\mathbf{I}_t \in \mathbb{R}^{3 \times H \times W}$ a model must predict a set of $K_t$ mask-label pairs $\mathcal{Y}_t = \{(\mathbf{m}_{t,i}, c_{t,i})\}_{i=1}^K$, where $\mathbf{m}_{t,i} \in \{0,1\}^{H \times W}$ is a binary mask and $c_{t,i} \in \{1,\ldots,C\}$ is a class label.
Importantly, for video segmentation, the model must not only produce accurate mask-label pairs for each frame, it should also maintain stable correspondences across frames. 
In particular, a pair $(\mathbf{m}_{t,i}, c_{t,i})$ for object $i$ at timestep $t$ should correspond to the same object as $(\mathbf{m}_{t-1,i}, c_{t-1,i})$ for object $i$ at timestep $t-1$, effectively tracking an object across time. 
This must be achieved in an \emph{online} manner: at timestep $t$, the predictions $\mathcal{Y}_t$ may only depend on the current frame $\mathbf{I}_t$ and the previously observed frames $\{\mathbf{I}_1, \dots, \mathbf{I}_{t-1}\}$.

\PAR{EoMT, VidEoMT, and PMT.}
EoMT~\cite{kerssies2025eomt} revisits image segmentation in the era of vision foundation models and shows that the  complex components of prior models~\cite{cheng2022mask2former,chen2023vitadapter} become largely redundant at increased model and pre-training scale. 
Therefore, it removes these components, and instead uses only a ViT encoder.
Like prior work~\cite{wang2021maxdeeplab,cheng2022mask2former,cavagnero2024pem}, EoMT operates on a set of $K$ learnable queries $\mathbf{Q}^\textrm{lrn} = \{\mathbf{q}^\textrm{lrn}_i \in \mathbb{R}^D\}_{i=1}^K$, where each query learns to represent a single object.
In EoMT, instead of using complex decoders, these queries are inserted directly into the ViT encoder after the first $L_1$ encoder layers, and the remaining $L_2$ layers process them together with the patch tokens as a single sequence. 
Finally, the model produces a segmentation mask and class label for each processed query with a lightweight head.
The simple design of EoMT leads to consistent efficiency gains over previous models. 

VidEoMT~\cite{norouzi2026videomt} extends this idea to online video segmentation by propagating queries across adjacent frames using a lightweight \emph{query fusion} layer. 
Specifically, the previous-frame output queries $\hat{\mathbf{Q}}_{t-1}$ are linearly projected and added to the learnable queries $\mathbf{Q}^\textrm{lrn}$ before being fed into the last $L_2$ layers of the encoder that processes current frame $\mathbf{I}_t$:
\begin{equation}
    \mathbf{Q}_{t}^{\mathcal{F}} = \texttt{Linear}\!\left(\hat{\mathbf{Q}}_{t-1}\right) + \mathbf{Q}^\textrm{lrn}.
\label{eq:query_fusion}
\end{equation}
The fused queries $\mathbf{Q}_{t}^{\mathcal{F}}$ replace $\mathbf{Q}^\textrm{lrn}$ in the last $L_2$ layers, so temporal information is propagated while the learnable queries support the detection of newly appearing objects.

Despite this streamlined design, both models require fine-tuning the full encoder, since the pre-trained attention layers must adapt to the injected queries.
PMT~\cite{cavagnero2026pmt} retains the same philosophy as EoMT and VidEoMT but moves query processing into a separate Segmenter-like~\cite{strudel2021segmenter} Transformer decoder. 
This decoupling allows the ViT encoder to remain frozen, making PMT compatible with multi-task deployment while preserving the simple design and efficiency of encoder-only architectures.\looseness=-1

\section{\ours{}} 
In this work, we take the state-of-the-art model PMT as our baseline and improve long-term modeling with two crucial improvements: $(i)$ We allow the model to adaptively retrieve and store object information in memory using a GRU (\cref{sec:gru_prop}). $(ii)$ We allow for training on long videos while limiting memory usage and preventing vanishing gradients using Truncated Query Propagation (\cref{sec:tqp}). The resulting model is called the \ours{} (\oursShort{}), and it is visualized in \cref{fig:arch}.

\label{sec:method}
\subsection{GRU-based Query Propagation}
\label{sec:gru_prop}
\PARbegin{Motivation.}
VidEoMT and PMT adopt the same query fusion strategy, defined in~\cref{eq:query_fusion}. 
At frame $t$, the input queries are obtained by linearly projecting the previous-frame output $\hat{\mathbf{Q}}_{t-1}$ and adding it to the learnable queries $\mathbf{Q}^\textrm{lrn}$. Although this strategy offers a simple and efficient mechanism for temporal propagation, it has a fundamental limitation: 
it does not allow the model to adaptively select the information that it wishes to keep, as the propagated queries are always just a simple sum of the learnable queries and the projected previous output.\looseness=-1

As a result, in case of long-term occlusions where an object disappears from the scene for a long time window, the query corresponding to this object is continuously updated with the learnable queries, while the per-frame encoder does not add any meaningful information to the propagated query because the object is not present in the frame. We expect that this causes the query to become diluted and lose information about the previous object, preventing it from being re-identified if it reappears after a long occlusion, making the model unsuitable for handling long-term occlusions.

\begin{figure*}[t]
\centering
\includegraphics[width=0.96\linewidth]{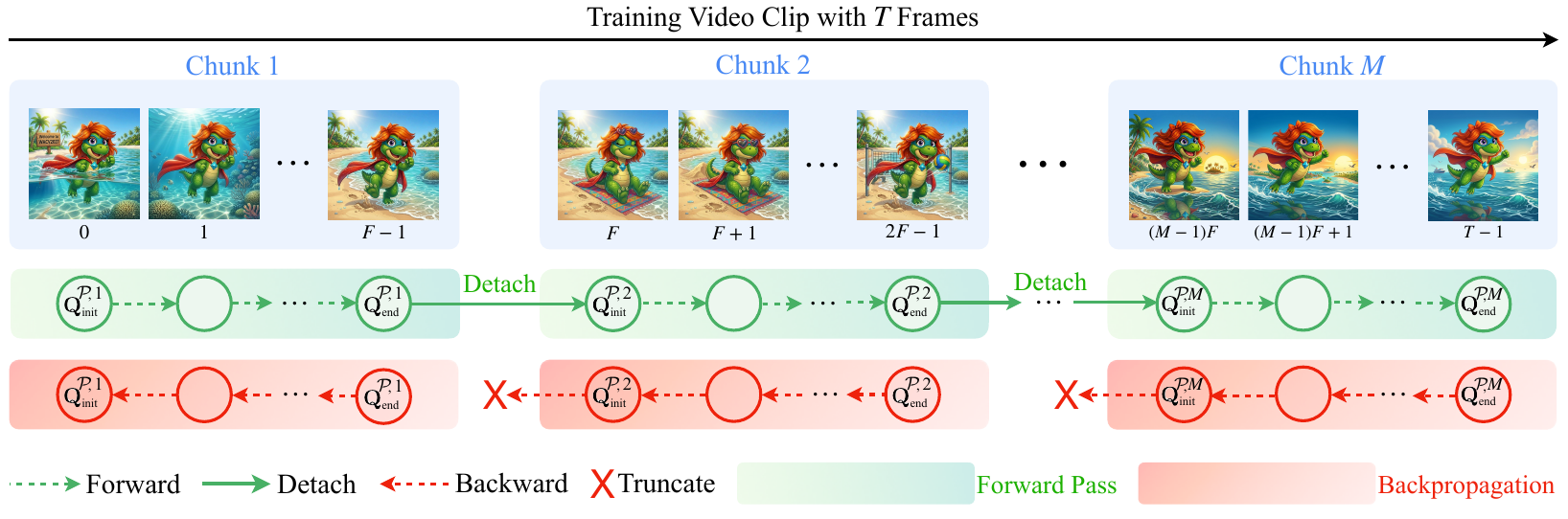}
\caption{\textbf{Truncated Query Propagation (\tqpshort{}).} A video of $T$ frames is partitioned into $M$ chunks of $F$ frames, and processed in order in a single iteration. For each chunk, we compute a
loss and backpropagate it to calculate the gradients. The optimizer is updated with the average, accumulated gradients after all chunks are consumed. Queries are detached and propagated across chunk boundaries, bounding peak memory to a single chunk and preventing the model from suffering from vanishing gradients caused by long-horizon recurrence.}
\label{fig:tqp}
\end{figure*}

\PAR{Method.}
To address this limitation, we introduce a lightweight memory mechanism for temporal propagation that can adaptively select which information it propagates across time.
Specifically, we replace the query fusion mechanism with a Gated Recurrent Unit (GRU)~\cite{cho2014gru}, a well-established recurrent architecture for modeling temporal dynamics.
In this GRU, the learnable queries act as the initial hidden state, and this hidden state is adaptively updated using the previous-frame output queries.
The updated hidden state is then fed into the current frame's decoder.

With this operation, the model has the freedom to be selective in which information is stored in the hidden state and thereby propagated across time.
As a result, when an object is no longer present, it can learn to keep information about this object to ensure that it can be re-identified when it reappears, allowing it to handle long-term occlusions. We empirically analyze this memory retention in \cref{sec:gru_memory_retention}.

\cref{fig:arch} shows how the GRU-based query propagation replaces the query propagation in the overall architecture.
Each individual frame is first fed into the frozen ViT and learned projection layers to obtain patch features $\mathbf{X}^l_t$.
Then, these patch features are fed into the Plain Mask Decoder (PMD) from PMT~\cite{cavagnero2026pmt} together with the propagated queries from the previous frame.
For the first frame, at $t=0$, PMD is fed the learnable queries $\mathbf{Q}^{\mathrm{lrn}}$ instead of the propagated queries, as propagated queries are not available yet:
\begin{equation}
    \mathbf{Q}^\mathcal{S}_{0} =
    \mathrm{PMD}\!\left(\mathbf{Q}^{\mathrm{lrn}},\, \mathbf{X}^l_0\right).
\label{eq:decoder_forward_t0}
\end{equation}
This yields segmentation queries $\mathbf{Q}^{\mathcal{S}}_{0}$ that are used to produce classification and mask predictions for frame $t=0$.

For the next frames, the propagated queries are produced using the GRU. 
Concretely, we use one GRU cell with shared parameters across all query slots.
Since no previous hidden state is available at $t=0$, we initialize the hidden state with the learnable queries:
\begin{equation}
    \mathbf{h}_{0} = \mathbf{Q}^{\mathrm{lrn}}, \qquad
    \mathbf{Q}^{\mathcal{P}}_{1} =
    \texttt{GRUCell}\!\left(\mathbf{Q}^{\mathcal{S}}_{0},\; \mathbf{h}_{0}\right).
\label{eq:gru_init_update_t0}
\end{equation}
Here, $\mathbf{Q}^{\mathcal{P}}_{1}$ denotes the propagated query representation passed to the next frame at $t=1$, which is equal to the updated hidden state $\mathbf{h}_{1}$.
For frames $t>0$, PMD takes as input the propagated query representation for the current frame, together with the corresponding lateral patch features.
\begin{equation}
    \mathbf{Q}^{\mathcal{S}}_{t} =
    \mathrm{PMD}\!\left(\mathbf{Q}^{\mathcal{P}}_{t},\, \mathbf{X}^l_t\right),
    \qquad t > 0.
\label{eq:decoder_forward_t}
\end{equation}

After decoding frame $t$, the resulting segmentation queries $\mathbf{Q}^{\mathcal{S}}_{t}$ are used for current-frame classification and mask prediction, and also for updating the recurrent state:
\begin{equation}
    \mathbf{Q}^{\mathcal{P}}_{t+1} =
    \texttt{GRUCell}\!\left(\mathbf{Q}^{\mathcal{S}}_{t},\; \mathbf{h}_{t}\right),
    \qquad t > 0.
\label{eq:gru_per_frame}
\end{equation}
Here, $\mathbf{Q}^{\mathcal{P}}_{t+1}$ denotes the propagated queries passed to PMD at timestep $t+1$, which are equal to the hidden state $\mathbf{h}_{t+1}$.

\subsection{Truncated Query Propagation (\tqpshort{})}
\label{sec:tqp}
\PARbegin{Motivation.}
After implementing the GRU-based propagation explained in \cref{sec:gru_prop}, we empirically observe that the segmentation accuracy and temporal consistency improves. 
However, we also find that identity preservation remains challenging and that identity switches still occur frequently.

When analyzing the cases for which the highest number of identity switches occur, we observe that errors typically occur for videos that contain many object instances, frequent object disappearances, and long disappearance spans (see \cref{sec:from_pmt_to_vmt}). 
Notably, the average disappearance span in these videos is about $2.6\times$ longer than the clip length used to train the model. 
Since the model is trained on short clips, it receives only limited temporal supervision during training, and the model is not exposed to enough disappearing and reappearing entities. 
Hence, in scenes where objects disappear for longer than the training horizon, the hidden state is insufficiently trained to preserve object identity across such interruptions. This can lead to identity switches at test time.

An intuitive next step is therefore to increase the training clip length, exposing the model to longer temporal dependencies. 
However, we find that naively training on longer clips is ineffective and introduces two problems. 
First, it causes a drop in accuracy, which we attribute to the harder optimization of recurrent propagation over long sequences, where gradients must pass through many GRU updates and gradually `vanish'. 
Second, it consistently increases memory consumption, often leading to out-of-memory errors.\looseness=-1

\PAR{Method.}
To address this, we apply the principle of Truncated Backpropagation Through Time~(TBPTT)~\cite{williams1990tbptt} to our GRU-based query propagation setting, which we refer to as \tqp{} (\tqpshort{}). 
In \tqpshort{}, the hidden state is propagated over long videos, but the gradient flow is truncated at the boundaries of short chunks.
This enables long-horizon supervision for the model without backpropagating through the full sequence, keeping training memory-efficient and improving optimization stability.

\tqpshort{}, visualized in \cref{fig:tqp}, partitions a training clip of $T$ frames into a sequence of $M=\lceil T/F \rceil$ chunks, denoted as $\mathcal{CH}=\{\mathrm{ch}_1,\mathrm{ch}_2,\ldots,\mathrm{ch}_M\}$, where each chunk contains at most $F$ frames. 
The chunks are processed in order within a single training iteration, and the optimizer updates the weights only once after all $M$ chunks of the video have been consumed. To optimize the model, \tqpshort{} leverages the average of the gradients computed for the individual chunks.

For each chunk $i$, we apply the forward and backward pass as usual, yielding a loss $\mathcal{L}^{(i)}$ and gradients to update the weights based on this loss.
Importantly, to enable the model to learn long-term temporal behavior, we allow information flow across chunks during the forward. Since the propagated queries are the GRU hidden state, i.e., $\mathbf{Q}^{\mathcal{P}}_{t} = \mathbf{h}_{t}$, we pass the final hidden state of chunk $i$ to the next chunk as the initial propagated query state, after detaching it from the computational graph:
\begin{equation}
    \mathbf{Q}^{\mathcal{P},\,i+1}_{\mathrm{init}}
    =
    \mathbf{h}^{\,i+1}_{\mathrm{init}}
    =
    \texttt{detach}\!\left(\mathbf{h}^{\,i}_{\mathrm{end}}\right).
\end{equation}

As a result, temporal information is preserved across chunks, while the computational graph remains bounded and full backpropagation through the entire sequence is avoided.
This enables long-horizon supervision for the GRU with bounded memory and more stable optimization.

\section{Experiments}
\label{sec:experiments}

\PARbegin{Datasets.} 
We evaluate \oursShort{} on six standard video segmentation benchmarks. 
For Video Instance Segmentation (VIS), we mainly focus on OVIS~\cite{qi2022occluded}, with challenging scenarios with heavy occlusions and crowded scenes, and YouTube-VIS 2022~\cite{yang2019video}, with long videos. We additionally evaluate on YouTube-VIS 2019 and 2021. For Video Panoptic Segmentation (VPS) we adopt VIPSeg~\cite{miao2022large} and for Video Semantic Segmentation (VSS) we use VSPW~\cite{miao2021vspw}.

\PARbegin{Implementation Details.}
Unless stated otherwise, we use a frozen ViT-L~\cite{dosovitskiy2021vit} encoder initialized with DINOv3~\cite{simeoni2025dinov3}.
Models are trained with the AdamW optimizer~\cite{loshchilov2019adamw} using mixed precision. 
By default, videos are processed in temporal chunks of $F=5$ frames, with an overall temporal window of $T=15$ frames, \ie, $M=3$ chunks.

For ground-truth matching, we follow standard practice: each object is matched to a query in the frame it first appears, and the assignment is kept across subsequent frames. 
For a fair comparison, we adopt the same batch size, learning rate, and learning rate scheduler as PMT~\cite{cavagnero2026pmt}, and the same video resolutions and number of training iterations as previous work~\cite{lee2025cavis,norouzi2026videomt,cavagnero2026pmt}; see \cref{sec:supp:impl_details} for details.
\PARbegin{Performance Metrics.} We evaluate our models using standard performance metrics for video segmentation. In particular, for VIS, we report Average Precision (AP) and Average Recall (AR)~\cite{yang2019video}. 
For VPS, we use Video Panoptic Quality (VPQ)~\cite{kim2020video} and Segmentation and Tracking Quality (STQ)~\cite{weber2021step}. 
For VSS, we adopt mean Intersection over Union (mIoU) and Video Consistency (mVC)~\cite{miao2021vspw}. 

We further evaluate the temporal consistency and identity preservation of our method with specialized tracking metrics.
We report IDF1~\cite{ristani2016performance}, Association Accuracy (AssA)~\cite{luiten2021hota}, mostly tracked (MT), mostly lost (ML)~\cite{dendorfer2021motchallenge}, and identity switches (IDS)~\cite{dendorfer2021motchallenge}. \cref{sec:supp:tracking_metrics} provides more details.

\PARbegin{Efficiency Metrics.}
For computational efficiency, we report both frames per second (FPS) and GFLOPs. 
FPS are measured as the average number of frames processed per second in the validation set at batch size 1, on a single NVIDIA H100 GPU with FlashAttention~v2 ~ \cite{dao2024flashattention2} and \texttt{torch.compile}~\cite{ansel2024pytorch2} enabled in default settings. 
GFLOPs are computed with \textit{fvcore}~\cite{meta2023fvcore} as the average over all validation images, where GFLOPs~$=$~FLOPs~$\times 10^{9}$.

\section{Results}
\label{sec:results}

\subsection{Main Results}
\label{sec:from_pmt_to_vmt}
In~\cref{tab:from_pmt_to_vmt}, we present a stepwise analysis of modifications from PMT~\cite{cavagnero2026pmt} toward our final \oursShort{} model, reporting mean AP over five runs. We choose PMT as our baseline because it achieves state-of-the-art accuracy at high FPS.

\PARbegin{GRU-based Query Propagation.}
In step~\tablestep{1}, we replace PMT's query fusion with our GRU-based query propagation. 
We find that this improves AP by $+2.4$ on the challenging OVIS dataset, with only a negligible impact on the number of parameters, GFLOPs and prediction speed. 
This suggests that allowing the model to adaptively select the information to store in memory allows it to propagate more useful information to the decoder that processes future frames, improving video segmentation performance.

Despite improved performance, we still empirically find that the model struggles to re-identify objects after long occlusions, causing erroneous \textit{identity switches}. 

To further explore this phenomenon, we compute some statistics for the 20 videos on which the model predictions from step~\tablestep{1} contain the \textit{most} and the \textit{least} identity switches.
The exact statistics are reported in \cref{sec:identity_consistency_analysis}.

We find that videos with the most identity switches contain about $3\times$ more objects per video, have $5\times$ more cases where objects are occluded (\ie, they disappear and reappear), and that the mean length of an occlusion is $2\times$ longer.
In fact, we find that the mean occlusion length is $13$ frames, which is $2.6\times$ longer than the training clips on which the model trains.
This suggests that the model should be trained on longer clips, to allow it to learn long-term modeling.

\begin{table}[t]
\centering
\scriptsize
\setlength{\tabcolsep}{1.2pt}
\renewcommand{\arraystretch}{1.05}

\resizebox{\linewidth}{!}{
\begin{tabular}{@{}lcccccccc@{}}
\toprule
Method & Step & GRU & TQP & $T_{\mathrm{train}}$
& Mean AP~$\uparrow$ & Params~$\downarrow$
& GFLOPs~$\downarrow$ & FPS~$\uparrow$ \\
\midrule

PMT~\cite{cavagnero2026pmt}
& \tablestep{0} & \xmark & \xmark & 5
& $51.8 \pm 0.3$ & 358M & 1014 & 97 \\

& \tablestep{1} & \cmark & \xmark & 5
& $54.2 \pm 0.4$ & 363M & 1015 & 95 \\

& \tablestep{2} & \cmark & \xmark & 10
& $52.3 \pm 0.3$ & 363M & 1015 & 95 \\

& \tablestep{3} & \cmark & \cmark & 10
& $56.1 \pm 0.3$ & 363M & 1015 & 95 \\

\textbf{\oursShort{} (Ours)}
& \tablestep{4} & \cmark & \cmark & 15
& $\mathbf{56.4 \pm 0.4}$ & 363M & 1015 & 95 \\

\bottomrule
\end{tabular}
}

\caption{\textbf{Stepwise modifications from PMT to \oursShort{} on OVIS
\textit{val}~\cite{qi2022occluded}.}
Mean AP $\pm$ standard deviation are reported over five independent runs.
$T_{\mathrm{train}}$ is the training-clip length in frames.}
\label{tab:from_pmt_to_vmt}
\vspace{-1pt}
\end{table}

\begin{table*}[t!]
    \centering
    \scriptsize
    \renewcommand{\tabcolsep}{3pt}
    \begin{tabularx}{\linewidth}
    {llcc c YYYYY c YYYYY}
    \toprule
    \multirow{2}[2]{*}{Method} &
    \multirow{2}[2]{*}{Backbone} &
    \multirow{2}[2]{*}{Pre-training} &
    \multirow{2}[2]{*}{Encoder} &&
    \multicolumn{5}{c}{OVIS \textit{val}~\cite{qi2022occluded}} &&
    \multicolumn{5}{c}{YouTube-VIS 2022 \textit{val}~\cite{yang2019video}} \\
    \cmidrule{6-10} \cmidrule{12-16}
    &&&&&
    AP & AP\textsubscript{75} & AR\textsubscript{10} & GFLOPs & FPS &&
    AP$^{\text{L}}$ & AP$^{\text{L}}_{\text{75}}$ & AR$^{\text{L}}_{\text{10}}$ & GFLOPs & FPS \\
    \midrule

    DVIS++~\cite{zhang2025dvis++} & ViT-Adapter-L~\cite{chen2023vitadapter} & DINOv2 & \twemoji{fire} &&
    49.6 & 55.0 & 54.6 & 868 & 17 &&
    37.5 & 39.4 & 43.5 & 820 & 18
    \\

    CAVIS~\cite{lee2025cavis} & ViT-Adapter-L~\cite{chen2023vitadapter} & DINOv2 & \twemoji{fire} &&
    53.2 & 59.1 & 58.2 & 863 & 15 &&
    39.5 & 40.5 & 44.9 & 815 & 15
    \\

    DVIS-DAQ~\cite{zhou2024dvisdaq}$^\dagger$ & ViT-Adapter-L~\cite{chen2023vitadapter} & DINOv2 & \twemoji{fire} &&
    54.3 & 60.2 & 59.8 & 1173 & 8 &&
    42.0 & 43.0 & 48.4 & 826 & 10
    \\

    LOMM~\cite{lee2025lomm} & ViT-Adapter-L~\cite{chen2023vitadapter} & DINOv2 & \twemoji{fire} &&
    51.7 & 57.5 & 56.2 & 899 & 12 &&
    48.2 & \underline{53.2} & 52.6 & 842 & 12
    \\

    VidEoMT~\cite{norouzi2026videomt}$^\dagger$ & ViT-L~\cite{dosovitskiy2021vit} & DINOv2 & \twemoji{fire} &&
    52.5 & 57.2 & 57.5 & 934 & 115 &&
    42.6 & 46.1 & 48.1 & 557 & 161
    \\
    
    VidEoMT~\cite{norouzi2026videomt}$^\dagger$ & ViT-L~\cite{dosovitskiy2021vit} & DINOv3 & \twemoji{fire} &&
    51.9 & 57.3 & 57.4 & 934 & 104 &&
    42.8 & 44.5 & 50.3 & 557 & 137
    \\

    \midrule

    CAVIS~\cite{lee2025cavis}$^\dagger$ & ViT-Adapter-L~\cite{chen2023vitadapter} & DINOv2 & \twemoji{2744} &&
    53.1 & 58.8 & 58.5 & 1168 & 10 &&
    41.4 & 41.0 & 47.4 & 815 & 15
    \\

    PMT~\cite{cavagnero2026pmt}$^\dagger$ & ViT-L~\cite{dosovitskiy2021vit} & DINOv2 & \twemoji{2744} &&
    51.8 & 57.7 & 56.0 & 1014 & 99 &&
    41.6 & 45.6 & 45.3 & 617 & 129
    \\

    \textbf{\oursShort{}} (Ours)$^\dagger$ & ViT-L~\cite{dosovitskiy2021vit} & DINOv2 & \twemoji{2744} &&
    \underline{55.6} & \underline{61.6} & \underline{60.5} & 1015 & 97 &&
    \underline{48.4} & 51.1 & 52.1 & 618 & 127
    \\

    \midrule

    CAVIS~\cite{lee2025cavis}$^\dagger$ & ViT-Adapter-L~\cite{chen2023vitadapter} & DINOv3 & \twemoji{2744} &&
    53.4 & 59.3 & 58.3 & 1168 & 9 &&
    42.2 & 37.6 & 49.6 & 815 & 13
    \\

    PMT~\cite{cavagnero2026pmt}$^\dagger$ & ViT-L~\cite{dosovitskiy2021vit} & DINOv3 & \twemoji{2744} &&
    52.0 & 56.0 & 57.7 & 1014 & 97 &&
    45.8 & 46.6 & \underline{53.2} & 617 & 123
    \\

    \textbf{\oursShort{}} (Ours)$^\dagger$ & ViT-L~\cite{dosovitskiy2021vit} & DINOv3 & \twemoji{2744} &&
    \textbf{56.7} & \textbf{61.8} & \textbf{61.6} & 1015 & 95 &&
    \textbf{51.5} & \textbf{58.3} & \textbf{55.4} & 618 & 121
    \\

    \bottomrule
    \end{tabularx}

    \caption{\textbf{\oursShort{} for VIS on OVIS and YouTube-VIS 2022~\cite{qi2022occluded,yang2019video}.} $^\dagger$Input resolution of 544 shortest image side for OVIS.
    }

    \label{tab:sota_comparison_ovis_ytvis22}
\end{table*}

\PARbegin{Training with Longer Clips.}
Therefore, in step~\tablestep{2}, we increase the training horizon  from $T_{\text{train}}=5$ to $T_{\text{train}}=10$ frames. 
We use $T_{\text{train}}=10$ as the longest feasible setting, since longer clips lead to out-of-memory errors. 
Interestingly, training with this longer horizon reduces the AP by $1.9$ points compared to step~\tablestep{1}.
This indicates that naively increasing the training clip does not automatically improve performance.
We expect that this happens because gradients `vanish' when optimizing the recurrent model on long sequences, harming the optimization process.
We confirm this through a gradient-flow analysis in \cref{sec:gradient_flow_analysis}.

\PARbegin{\tqpshort{} Training Strategy.}
To address this issue, we apply our \tqpshort{} training strategy in step~\tablestep{3}.
\tqpshort{} yields a significant $+3.8$ AP boost compared to naive longer-clip training, while maintaining inference efficiency.
When using even longer clips in the final step~\tablestep{4}, performance is boosted slightly further.
This demonstrates that \tqpshort{}'s chunked training allows the model to fully benefit from long-clip training without being impacted by vanishing gradients or increased memory usage.
Overall, the results show that \oursShort{}'s use of GRU-based propagation and long-clip training with \tqpshort{} allow it to outperform the state-of-the-art PMT baseline by a significant $+4.6$ AP while preserving simplicity and efficiency, showcasing its effectiveness.

\subsection{Comparison with State-of-the-Art Models}
\PARbegin{Video Instance Segmentation (VIS).}
In \cref{tab:sota_comparison_ovis_ytvis22}, we compare \oursShort{} to existing state-of-the-art models on the challenging OVIS~\cite{qi2022occluded} and YouTube-VIS 2022~\cite{yang2019video} benchmarks.
We compare to two categories of models: those that fine-tune the encoder and those that keep the encoder frozen. 
In \oursShort{}, we decide to keep the encoder frozen like in PMT~\cite{cavagnero2026pmt} because (i) this allows the encoder to be reused for other tasks, and (ii) it simply performs better, as we show in more detail in \cref{sec:effect_of_fine-tuning}.\looseness=-1

Comparing to existing models that also keep the encoder frozen, \oursShort{} obtains a much higher accuracy, with $+3.3$ AP on OVIS and $+9.3$ AP on YouTube-VIS 2022 compared to CAVIS~\cite{lee2025cavis} with DINOv3, while being much faster. 
As already demonstrated in \cref{sec:from_pmt_to_vmt}, \oursShort{} also performs considerably better than PMT~\cite{cavagnero2026pmt}, while preserving its efficiency.

\oursShort{} also significantly outperforms methods that \textit{do} finetune the encoder.
Specifically, it beats current state-of-the-art method DVIS-DAQ~\cite{zhou2024dvisdaq} on OVIS by $+2.4$ AP and LOMM~\cite{lee2025lomm} on YouTube-VIS 2022 by $+3.3$ AP, while being over 10$\times$ faster than both of them.
When using the same DINOv2 encoder, \oursShort{} performs comparably to LOMM at an AP of 48.4 and 48.2, respectively, while still being much faster.
Compared to the highly efficient method VidEoMT~\cite{norouzi2026videomt}, \oursShort{} obtains a considerably higher accuracy, at $+4.2$ AP for OVIS and $+8.7$ from YouTube-VIS 2022, while being only slightly less fast.

In \cref{sec:supp:ytvis1921_comparison}, we demonstrate that similar trends hold on the YouTube-VIS 2019 and 2021 datasets, albeit with smaller absolute differences due to the lower complexity of these datasets. Overall, these results demonstrate the strength and effectiveness of \oursShort{} and its components for video segmentation on complex and long videos.

\begin{table}[t!]
    \centering
    \scriptsize
    \renewcommand{\tabcolsep}{2.5pt}

    \begin{tabular*}{\linewidth}{@{\extracolsep{\fill}}lccccc}
    \toprule
    Method & IDF1 (\%) $\uparrow$ & AssA (\%) $\uparrow$ & MT (\%) $\uparrow$ & ML (\%) $\downarrow$ & Total IDS $\downarrow$ \\
    \midrule
    
    CAVIS~\cite{lee2025cavis} &
    79.2 & 74.7 & 76.3 & 6.1 & 2683 \\
    
    VidEoMT~\cite{norouzi2026videomt} &
    78.7 & 73.3 & 75.5 & 6.2 & 2801 \\
    
    PMT~\cite{cavagnero2026pmt} &
    78.9 & 73.3 & 75.7 & 6.3 & 2755 \\

    \textbf{\oursShort{}} (Ours) &
    81.8 & 78.4 & 78.8 & 5.5 & 2001 \\
    
    \bottomrule
    \end{tabular*}
    \caption{\textbf{Tracking quality on OVIS \textit{val}~\cite{qi2022occluded}.} We compare~\oursShort{} with existing methods using specialized tracking metrics.}
    \label{tab:tracking_quality_comparison}
    \vspace{-3pt}
\end{table}

\PARbegin{Tracking Quality.}
\label{sec:tracking_quality}
\cref{tab:tracking_quality_comparison} further analyzes the tracking performance of \oursShort{} on OVIS using specialized tracking metrics. 
\oursShort{} achieves the best performance across all reported metrics. 
Most notably, compared to PMT, it reduces the number of identity switches by about $27\%$. 
This shows that \oursShort{} reduces identity errors, as intended.
We also provide qualitative results in \cref{sec:supp:qual_results} to further illustrate the tracking quality of \oursShort{}.

\begin{table}[t]
    \centering
    \scriptsize
    \renewcommand{\tabcolsep}{2pt}
    \begin{tabularx}{\linewidth}{llcc c cYYY}
    \toprule
    \multirow{2}[2]{*}{Method} &
    \multirow{2}[2]{*}{Backbone} &
    \multirow{2}[2]{*}{PT} &
    \multirow{2}[2]{*}{Encoder} && 
    \multicolumn{4}{c}{VIPSeg \textit{val}~\cite{miao2022large}} \\
    \cmidrule{6-9}
    &&&&&
    VPQ & STQ & GFLOPs & FPS \\
    \midrule

    DVIS++~\cite{zhang2025dvis++} & ViT-Adapter-L~\cite{chen2023vitadapter} & D2 & \twemoji{fire} &
    & 56.0 & 49.8 & 2290 & 13 \\

    DVIS-DAQ~\cite{zhou2024dvisdaq} & ViT-Adapter-L~\cite{chen2023vitadapter} & D2 & \twemoji{fire} &
    & \underline{57.4} & \underline{52.0} & 2315 & 4 \\

    CAVIS~\cite{lee2025cavis} & ViT-Adapter-L~\cite{chen2023vitadapter} & D2 & \twemoji{fire} &
    & 56.9 & 51.0 & 2612 & 10 \\

    VidEoMT~\cite{norouzi2026videomt} & ViT-L~\cite{dosovitskiy2021vit} & D2 & \twemoji{fire} &
    & 55.2 & 48.9 & 1897 & 75 \\

    VidEoMT~\cite{norouzi2026videomt} & ViT-L~\cite{dosovitskiy2021vit} & D3 & \twemoji{fire} &
    & 55.1 & 48.1 & 1897 & 71 \\

    \midrule

    CAVIS~\cite{lee2025cavis} & ViT-Adapter-L~\cite{chen2023vitadapter} & D2 & \twemoji{2744} &
    & 56.4 & 49.0 & 2612 & 10 \\

    PMT~\cite{cavagnero2026pmt} & ViT-L~\cite{dosovitskiy2021vit} & D2 & \twemoji{2744} &
    & 55.3 & 48.2 & 2037 & 60 \\

    \textbf{\oursShort{}} (Ours) & ViT-L~\cite{dosovitskiy2021vit} & D2 & \twemoji{2744} &
    & 56.6 & 51.3 & 2038 & 59 \\

    \midrule

    CAVIS~\cite{lee2025cavis} & ViT-Adapter-L~\cite{chen2023vitadapter} & D3 & \twemoji{2744} &
    & 56.8 & 51.2 & 2612 & 9 \\

    PMT~\cite{cavagnero2026pmt} & ViT-L~\cite{dosovitskiy2021vit} & D3 & \twemoji{2744} &
    & 55.5 & 49.2 & 2037 & 58 \\
    
    \textbf{\oursShort{}} (Ours) & ViT-L~\cite{dosovitskiy2021vit} & D3 & \twemoji{2744} &
    & \textbf{60.3} & \textbf{53.4} & 2038 & 57 \\

    \bottomrule
    \end{tabularx}
    \caption{\textbf{\oursShort{} for VPS on VIPSeg~\cite{miao2022large}.} \textbf{PT}: Pre-training. \textbf{D2}: DINOv2~\cite{oquab2023dinov2}. \textbf{D3}: DINOv3~\cite{simeoni2025dinov3}.
}
    \label{tab:sota_comparison_vipseg}
    \vspace{-4pt}
\end{table}

\PARbegin{Video Panoptic Segmentation (VPS).}
\oursShort{} also obtains new state-of-the-art results for VPS on the VIPSeg~\cite{miao2022large} benchmark, as demonstrated in \cref{tab:sota_comparison_vipseg}.
It significantly outperforms both PMT and CAVIS that also use a frozen encoder, while being much faster than CAVIS.
Compared to existing state-of-the-art model DVIS-DAQ, which finetunes the encoder, \oursShort{} improves performance by $+2.9$ VPQ when using DINOv3 pretraining.
When using DINOv2, the models perform more similarly at a $0.8$ VPQ difference, but \oursShort{} is over 10$\times$ faster and allows the encoder to be reused, making it more practically useful.
These results confirm \oursShort{}'s effectiveness in obtaining state-of-the-art accuracy for video segmentation while maintaining high FPS.\looseness=-1

\begin{table}[t]
    \centering
    \scriptsize
    \renewcommand{\tabcolsep}{2pt}
    \begin{tabularx}{\linewidth}{llcc c cccc}
    \toprule
    \multirow{2}[2]{*}{Method} &
    \multirow{2}[2]{*}{Backbone} &
    \multirow{2}[2]{*}{PT} &
    \multirow{2}[2]{*}{Encoder} && 
    \multicolumn{4}{c}{VSPW \textit{val}~\cite{miao2021vspw}} \\
    \cmidrule{6-9}
    &&&&&
    mVC\textsubscript{16} & mIoU & GFLOPs & FPS \\
    \midrule

    DVIS++~\cite{zhang2025dvis++} & ViT-Adapter-L~\cite{chen2023vitadapter} & D2 & \twemoji{fire} &
    & 94.2 & 62.8 & 2290 & 13 \\

    VidEoMT~\cite{norouzi2026videomt} & ViT-L~\cite{dosovitskiy2021vit} & D2 & \twemoji{fire} &
    & 95.0 & 64.9 & 1909 & 73 \\

    VidEoMT~\cite{norouzi2026videomt} & ViT-L~\cite{dosovitskiy2021vit} & D3 & \twemoji{fire} &
    & 94.4 & 64.0 & 1909 & 71 \\

    \midrule

    PMT~\cite{cavagnero2026pmt} & ViT-L~\cite{dosovitskiy2021vit} & D2 & \twemoji{2744} &
    & 94.6 & 64.3 & 2049 & 60 \\

    \textbf{\oursShort{}} (Ours) & ViT-L~\cite{dosovitskiy2021vit} & D2 & \twemoji{2744} &
    & \underline{95.2} & 65.4 & 2050 & 59 \\

    \midrule

    PMT~\cite{cavagnero2026pmt} & ViT-L~\cite{dosovitskiy2021vit} & D3 & \twemoji{2744} &
    & 94.9 & \underline{65.7} & 2049 & 58 \\

    \textbf{\oursShort{}} (Ours) & ViT-L~\cite{dosovitskiy2021vit} & D3 & \twemoji{2744} &
    & \textbf{95.3} & \textbf{66.4} & 2050 & 57 \\

    \bottomrule
    \end{tabularx}
    \caption{\textbf{\oursShort{} for VSS on VSPW~\cite{miao2021vspw}.} \textbf{PT}: Pre-training. \textbf{D2}: DINOv2~\cite{oquab2023dinov2}. \textbf{D3}: DINOv3~\cite{simeoni2025dinov3}.
}
    \label{tab:sota_comparison_vspw}
\end{table}

\PARbegin{Video Semantic Segmentation (VSS).}
The strength of \oursShort{} is further demonstrated for the VSS task on the VSPW~\cite{miao2021vspw} benchmark in \cref{tab:sota_comparison_vspw}.
Even with a frozen encoder, \oursShort{} surpasses VidEoMT~\cite{norouzi2026videomt} by $+0.9$ mVC and $+2.4$ mIoU with DINOv3, while having only slightly lower FPS. 
Compared to PMT~\cite{cavagnero2026pmt}, which operates under the same frozen-encoder setting and virtually identical FPS, \oursShort{} gains $+0.4$ mVC and $+0.7$ mIoU in combination with DINOv3.
In short, \oursShort{} also sets a new state of the art on VSPW while preserving high efficiency of current models.

\subsection{Ablations}
We report the main ablation studies in this section, with additional results provided in \cref{sec:supp:add_ablations}.

\PARbegin{Choice of Memory Mechanism.}
\oursShort{} uses a GRU cell~\cite{cho2014gru} as the memory mechanism that keeps track of the queries representing the tracked objects.
To assess the effectiveness of this design choice, we compare it to alternative memory mechanisms in \cref{tab:ablation_memory_ovis}.
We consider recurrent units, including LSTM~\cite{hochreiter1997long}, GRU~\cite{cho2014gru}, and RVM's DeepGRU~\cite{zoran2025rvm};  a structured state-space model, S5~\cite{smith2023simplified}, designed for efficient long-range sequence modeling; and an explicit memory-bank design with cross-attention, following GenVIS~\cite{heo2023generalized}. 

Among these choices, the GRU provides the best overall trade-off, achieving the highest AP while remaining among the fastest variants. 
In particular, it improves over LSTM by $+0.7$ AP at slightly higher FPS, and over the memory-bank baseline by a larger margin of $+2.2$ AP.
We expect that the the simple GRU works better than the alternatives because they add unnecessary complexity that complicates the model's optimization process.
For the task of query propagation, it appears to be sufficient to equip the model with a simple mechanism that it can use to adaptively select which information it stores in memory and propagates, without requiring complex operations.

\begin{table}[t!]
    \centering
    \scriptsize
    \renewcommand{\tabcolsep}{2.5pt}
    \begin{tabularx}{\linewidth}{l YYYYY}
    \toprule
    Memory &
    AP & AP\textsubscript{75} & AR\textsubscript{10} & GFLOPs & FPS \\
    \midrule

    LSTM~\cite{hochreiter1997long} &
    56.0 & 61.3 & 60.9 & 1016 & 93 \\

    S5~\cite{smith2023simplified} &
    55.8 & 61.8 & 60.5 & 1014 & 96 \\

    DeepGRU~\cite{zoran2025rvm} &
    56.5 & 61.8 & 61.5 & 1019 & 91 \\

    Memory Bank~\cite{heo2023generalized} &
    54.5 & 59.8 & 59.9 & 1017 & 86 \\ \midrule

    \textbf{GRU}~\cite{cho2014gru} &
    56.7 & 61.8 & 61.6 & 1015 & 95 \\

    \bottomrule
    \end{tabularx}
    \caption{\textbf{Effect of memory mechanism on OVIS \textit{val}~\cite{qi2022occluded}.} We compare various memory mechanisms for temporal modeling.
    }
    \label{tab:ablation_memory_ovis}
\end{table}

\begin{table}[t!]
    \centering
    \scriptsize
    \renewcommand{\tabcolsep}{2.5pt}
    \begin{tabularx}{\linewidth}{ccc YYYYY}
    \toprule
    $T_\textrm{train}$ & GRU & TQP &
    AP & AP\textsubscript{75} & AR\textsubscript{10} & GFLOPs & FPS \\
    \midrule

    5 & \xmark & \xmark &
    52.0 & 56.0 & 57.7 & 1014 & 97 \\

    5 & \cmark & \xmark &
    54.7 & 60.6 & 59.7 & 1015 & 95 \\

    \midrule

    10 & \xmark & \xmark &
    51.1 & 56.0 & 56.7 & 1014 & 97 \\

    10 & \cmark & \xmark &
    52.6 & 56.6 & 58.4 & 1015 & 95 \\

    10 & \xmark & \cmark &
    54.5 & 59.6 & 59.9 & 1014 & 97 \\

    10 & \cmark & \cmark &
    56.5 & 61.5 & 61.5 & 1015 & 95 \\

    \midrule

    15 & \xmark & \xmark &
    \textcolor{lightgray}{OOM} & \textcolor{lightgray}{OOM} & \textcolor{lightgray}{OOM} & \textcolor{lightgray}{OOM} & \textcolor{lightgray}{OOM} \\

    15 & \cmark & \xmark &
    \textcolor{lightgray}{OOM} & \textcolor{lightgray}{OOM} & \textcolor{lightgray}{OOM} & \textcolor{lightgray}{OOM} & \textcolor{lightgray}{OOM} \\

    15 & \xmark & \cmark &
    54.6 & 60.1 & 59.5 & 1014 & 97 \\

    15 & \cmark & \cmark &
    56.7 & 61.8 & 61.6 & 1015 & 95 \\

    \midrule

    20 & \xmark & \xmark &
    \textcolor{lightgray}{OOM} & \textcolor{lightgray}{OOM} & \textcolor{lightgray}{OOM} & \textcolor{lightgray}{OOM} & \textcolor{lightgray}{OOM} \\

    20 & \cmark & \xmark &
    \textcolor{lightgray}{OOM} & \textcolor{lightgray}{OOM} & \textcolor{lightgray}{OOM} & \textcolor{lightgray}{OOM} & \textcolor{lightgray}{OOM} \\

    20 & \xmark & \cmark &
    54.6 & 60.2 & 59.3 & 1014 & 97 \\

    20 & \cmark & \cmark &
    56.8 & 62.2 & 61.9 & 1015 & 95 \\

    \bottomrule
    \end{tabularx}
    \caption{\textbf{Effect of GRU and \tqpshort{} on OVIS \textit{val}~\cite{qi2022occluded}.}
    We evaluate the effects of GRU-based query propagation and \tqpshort{} across different training clip lengths $T_\textrm{train}$. All results are measured using 8 NVIDIA H100 (94GB) with batch size 8 (one clip per GPU) and multi-scale resolution (320--640px shortest side, capped at 768).}
    \label{tab:ablation_num_frames_gru_TQP_ovis}
    \vspace{-5pt}
\end{table}

\PARbegin{Effect of GRU and \tqpshort{}.}
\label{sec:ablation_gru_tqp}
\cref{tab:ablation_num_frames_gru_TQP_ovis} studies the individual and joint effects of GRU-based query propagation and \tqpshort{} for increasing training clip length. 
As already shown in \cref{tab:from_pmt_to_vmt}, GRU-based propagation boosts performance when training on short clips but causes a performance drop when naively training on long clips, which is solved by adopting \tqpshort{}'s chunked training mechanism.

The results in \cref{tab:ablation_num_frames_gru_TQP_ovis} additionally show that naive training on long clips \emph{without} GRU-based propagation does not work well either, with $51.1$ \vs $52.6$ AP on 10 frames. 
This suggests that the vanishing gradient problem is not unique to the GRU mechanism, but also occurs for the default recurrent query fusion mechanism used by the PMT baseline.

Similarly, the results also show that a model \emph{with} \tqpshort{} but \emph{without} GRU propagation does not perform as a model that uses both, with $54.5$ \vs $56.5$ AP on 10 frames.
This indicates that both TQP and GRU are critical in achieving a good performance for long and complex video segmentation, and that both these contributions are complementary.

Finally, we find that training without \tqpshort{} results in out-of-memory errors for clips of 15 frames or longer, whereas \tqpshort{} enables training at these clip lengths. However, increasing $T_{\mathrm{train}}$ from 15 to 20 frames yields only a modest gain of $+0.1$ AP. This is consistent with our analysis in \cref{sec:from_pmt_to_vmt}, which shows that the average occlusion length in OVIS is 13 frames. Since $T_{\mathrm{train}}=15$ already covers the average occlusion. We therefore expect our long-term training strategy to provide greater benefits on datasets with longer occlusions, where longer training clips would expose the model to longer occlusion events. In this work, we use $T_\textrm{train}=15$ frames by default, as it provides a good trade-off between performance and training time.

\section{Conclusion}
\label{sec:conclusion}

In this paper, we have addressed the challenge of long-term object tracking for video segmentation tasks.
Existing video segmentation methods struggle with long-term tracking because $(i)$ they do not have a mechanism that can adaptively store relevant information about tracked objects, and $(ii)$ they cannot be trained effectively on long videos.
We addressed these limitations by $(i)$ employing a lightweight GRU-memory that can adaptively propagate relevant information across time, and $(ii)$ introducing \tqp{} (\tqpshort{}), in which the model is trained on chunks of videos to enable stable and memory-efficient optimization on long sequences.
The resulting model, the \ours{}~(\oursShort{}), outperforms state-of-the-art methods across all video segmentation tasks while being over 10$\times$ faster. At the same time, it is consistently more accurate than the efficient PMT baseline at comparable speed.

\subsubsection*{Acknowledgments}
This work was partly funded by the Cynergy4MIE project, supported by the Chips Joint Undertaking and its members, including top-up funding from National Authorities under Grant Agreement No. 101140226, the BMFTR project WestAI (grant no. 16IS22094D), and the EU project JUPITER AI Factory (grant no. 101250682). The experiments utilized the Dutch national infrastructure, supported by the SURF Cooperative under grant nos. EINF-14337 and EINF-17956 and funded by the Dutch Research Council (NWO), computing resources granted by RWTH Aachen under project seg4video, and resources provided by the Gauss Centre for Supercomputing e.V. through the John von Neumann Institute for Computing (NIC) on the GCS supercomputers JUWELS / JUPITER at the Jülich Supercomputing Centre.

{
    \small
    \bibliographystyle{ieeenat_fullname}
    \bibliography{main}

@String(PAMI = {IEEE Trans. Pattern Anal. Mach. Intell.})

@String(IJCV = {Int. J. Comput. Vis.})

@String(CVPR= {IEEE Conf. Comput. Vis. Pattern Recog.})

@String(ICCV= {Int. Conf. Comput. Vis.})

@String(ECCV= {Eur. Conf. Comput. Vis.})

@String(ICLR = {Int. Conf. Learn. Represent.})

@String(CVPRW= {IEEE Conf. Comput. Vis. Pattern Recog. Worksh.})

@String(TMLR = {Transactions on Machine Learning Research})

@String(EMNLP = {Empirical Methods in Natural Language Processing})

@String(PAMI  = {IEEE TPAMI})

@String(IJCV  = {IJCV})

@String(CVPR  = {CVPR})

@String(ICCV  = {ICCV})

@String(ECCV  = {ECCV})

@String(NeurIPS  = {NeurIPS})

@String(ICLR  = {ICLR})

@String(CVPRW= {CVPRW})

@String(ASPLOS = {ASPLOS})

@inproceedings{kerssies2025eomt,
    title={{Your ViT is Secretly an Image Segmentation Model}},
    author={Kerssies, Tommie and Cavagnero, Niccol\`{o} and Hermans, Alexander and Norouzi, Narges and Averta, Giuseppe and Leibe, Bastian and Dubbelman, Gijs and de Geus, Daan},
    booktitle=CVPR,
    year=2025
}

@inproceedings{norouzi2026videomt,
    title={{VidEoMT: Your ViT is Secretly Also a Video Segmentation Model}},
    author={Norouzi, Narges and Zulfikar, Idil and Cavagnero, Niccol\`{o} and Kerssies, Tommie and Leibe, Bastian and Dubbelman, Gijs and {de Geus}, Daan},
    booktitle=CVPR,
    year=2026
}

@inproceedings{cavagnero2026pmt,
    title={{PMT: Plain Mask Transformer for Image and Video Segmentation with Frozen Vision Encoders}},
    author={Cavagnero, Niccol\`{o} and Norouzi, Narges and Dubbelman, Gijs and {de Geus}, Daan},
    booktitle=CVPRW,
    year=2026
}

@inproceedings{yang2019video,
    title={{Video Instance Segmentation}},
    author={Yang, Linjie and Fan, Yuchen and Xu, Ning and Yang, Ding and Yue, Dong and Liang, Jianchao and Huang, Thomas and Huang, Humphrey},
    booktitle=ICCV,
    year=2019
}

@inproceedings{nilsson2018semantic,
    title={{Semantic Video Segmentation by Gated Recurrent Flow Propagation}},
    author={Nilsson, David and Sminchisescu, Cristian},
    booktitle=CVPR,
    year=2018
}

@inproceedings{kim2020video,
    title={{Video Panoptic Segmentation}},
    author={Kim, Dahun and Woo, Sanghyun and Lee, Joon-Young and Kweon, In So},
    booktitle=CVPR,
    year=2020
}

@inproceedings{wu2022defense,
    title={{In Defense of Online Models for Video Instance Segmentation}},
    author={Wu, Junfeng and Liu, Qihao and Jiang, Yi and Bai, Song and Yuille, Alan and Bai, Xiang},
    booktitle=ECCV,
    year=2022
}

@inproceedings{huang2022minvis,
    title={{MinVIS: A Minimal Video Instance Segmentation Framework without Video-based Training}},
    author={Huang, De-An and Yu, Zhiding and Anandkumar, Anima},
    booktitle=NeurIPS,
    year=2022
}

@inproceedings{heo2023generalized,
    title={{A Generalized Framework for Video Instance Segmentation}},
    author={Heo, Miran and Hwang, Sukjun and Hyun, Jeongseok and Kim, Hanjung and Oh, Seoung Wug and Lee, Joon-Young and Kim, Seon Joo},
    booktitle=CVPR,
    year=2023
}

@inproceedings{zhang2023dvis,
    title={{DVIS: Decoupled Video Instance Segmentation Framework}},
    author={Zhang, Tao and Tian, Xingye and Wu, Yu and Ji, Shunping and Wang, Xuebo and Zhang, Yuan and Wan, Pengfei},
    booktitle=ICCV,
    year=2023
}

@inproceedings{lee2025cavis,
    title={{Context-Aware Video Instance Segmentation}},
    author={Lee, Seunghun and Seo, Jiwan and Han, Kiljoon and Choi, Minwoo and Im, Sunghoon},
    booktitle=ICCV,
    year=2025
}

@inproceedings{zhou2024dvisdaq,
    title={{Improving Video Segmentation via Dynamic Anchor Queries}},
    author={Zhou, Yikang and Zhang, Tao and Ji, Shunping and Yan, Shuicheng and Li, Xiangtai},
    booktitle=ECCV,
    year=2024
}

@article{zhang2025dvis++,
    title={{DVIS++: Improved Decoupled Framework for Universal Video Segmentation}},
    author={Zhang, Tao and Tian, Xingye and Zhou, Yikang and Ji, Shunping and Wang, Xuebo and Tao, Xin and Zhang, Yuan and Wan, Pengfei and Wang, Zhongyuan and Wu, Yu},
    journal=PAMI,
    year=2025
}

@inproceedings{lee2025lomm,
    title= {{LOMM: Latest Object Memory Management for Temporally Consistent Video Instance Segmentation}},
    author={Lee, Seunghun and Seo, Jiwan and Choi, Minwoo and Han, Kiljoon and Jeong, Jahoon and Durante, Zane and Adeli, Ehsan and Park, Sang Hyun and Im, Sunghoon},
    booktitle=ICCV,
    year=2025
}

@inproceedings{cheng2022xmem,
    title={{XMem}: Long-Term Video Object Segmentation with an Atkinson-Shiffrin Memory Model},
    author={Cheng, Ho Kei and Alexander G. Schwing},
    booktitle=ECCV,
    year=2022
}

@inproceedings{cheng2024putting,
  title={{Putting the Object Back into Video Object Segmentation}},
  author={Cheng, Ho Kei and Oh, Seoung Wug and Price, Brian and Lee, Joon-Young and Schwing, Alexander},
  booktitle=CVPR,
  year=2024
}

@inproceedings{carion2025sam3segmentconcepts,
    title={{SAM 3: Segment Anything with Concepts}},
    author={Nicolas Carion and Laura Gustafson and Yuan-Ting Hu and Shoubhik Debnath and Ronghang Hu and Didac Suris and Chaitanya Ryali and Kalyan Vasudev Alwala and Haitham Khedr and Andrew Huang and Jie Lei and Tengyu Ma and Baishan Guo and Arpit Kalla and Markus Marks and Joseph Greer and Meng Wang and Peize Sun and Roman Rädle and Triantafyllos Afouras and Effrosyni Mavroudi and Katherine Xu and Tsung-Han Wu and Yu Zhou and Liliane Momeni and Rishi Hazra and Shuangrui Ding and Sagar Vaze and Francois Porcher and Feng Li and Siyuan Li and Aishwarya Kamath and Ho Kei Cheng and Piotr Dollár and Nikhila Ravi and Kate Saenko and Pengchuan Zhang and Christoph Feichtenhofer},
    booktitle=ICLR,
    year=2026
}

@inproceedings{meinhardt2022trackformer,
    title={{TrackFormer: Multi-Object Tracking with Transformers}},
    author={Meinhardt, Tim and Kirillov, Alexander and Leal-Taixe, Laura and Feichtenhofer, Christoph},
    booktitle=CVPR,
    year=2022
}

@inproceedings{wu2006tracking,
    title={{Tracking of Multiple, Partially Occluded Humans based on Static Body Part Detection}},
    author={Wu, Bo and Nevatia, Ram},
    booktitle=CVPR,
    year=2006
}

@inproceedings{wang2021maxdeeplab,
  title={{MaX-DeepLab: End-to-End Panoptic Segmentation with Mask Transformers}},
  author={Wang, Huiyu and Zhu, Yukun and Adam, Hartwig and Yuille, Alan and Chen, Liang-Chieh},
  booktitle= CVPR,
  year=2021
}

@inproceedings{strudel2021segmenter,
    title={{Segmenter: Transformer for Semantic Segmentation}},
    author={Strudel, Robin and Garcia, Ricardo and Laptev, Ivan and Schmid, Cordelia},
    booktitle=ICCV,
    year=2021
}

@inproceedings{cheng2022mask2former,
    title={{Masked-attention Mask Transformer for Universal Image Segmentation}},
    author={Cheng, Bowen and Misra, Ishan and Schwing, Alexander G. and Kirillov, Alexander and Girdhar, Rohit},
    booktitle=CVPR,
    year=2022
}

@inproceedings{cavagnero2024pem,
    title={{PEM: Prototype-based Efficient MaskFormer for Image Segmentation}},
    author={Cavagnero, Niccol\`{o} and Rosi, Gabriele and Cuttano, Claudia and Pistilli, Francesca and Ciccone, Marco and Averta, Giuseppe and Cermelli, Fabio},
    booktitle=CVPR,
    year=2024
}

@article{qi2022occluded,
    title={{Occluded Video Instance Segmentation: A Benchmark}},
    author={Qi, Jiyang and Gao, Yan and Hu, Yao and Wang, Xinggang and Liu, Xiaoyu and Bai, Xiang and Belongie, Serge and Yuille, Alan and Torr, Philip HS and Bai, Song},
    journal=IJCV,
    year=2022
}

@inproceedings{miao2021vspw,
    title={{VSPW: A Large-scale Dataset for Video Scene Parsing in the Wild}},
    author={Miao, Jiaxu and Wei, Yunchao and Wu, Yu and Liang, Chen and Li, Guangrui and Yang, Yi},
    booktitle=CVPR,
    year=2021
}

@inproceedings{miao2022large,
    title={{Large-Scale Video Panoptic Segmentation in the Wild: A Benchmark}},
    author={Miao, Jiaxu and Wang, Xiaohan and Wu, Yu and Li, Wei and Zhang, Xu and Wei, Yunchao and Yang, Yi},
    booktitle=CVPR,
    year=2022
}

@inproceedings{weber2021step,
    title={{STEP: Segmenting and Tracking Every Pixel}},
    author={Weber, Mark and Xie, Jun and Collins, Maxwell and Zhu, Yukun and Voigtlaender, Paul and Adam, Hartwig and Green, Bradley and Geiger, Andreas and Leibe, Bastian and Cremers, Daniel and others},
    booktitle=NeurIPS,
    year=2021
}

@inproceedings{ristani2016performance,
    title={{Performance Measures and a Data Set for Multi-Target, Multi-Camera Tracking}},
    author={Ristani, Ergys and Solera, Francesco and Zou, Roger S. and Cucchiara, Rita and Tomasi, Carlo},
    booktitle=ECCV,
    year=2016
}

@article{luiten2021hota,
    title={{HOTA: A Higher Order Metric for Evaluating Multi-Object Tracking}},
    author={Luiten, Jonathon and Osep, Aljosa and Dendorfer, Patrick and Torr, Philip H. S. and Geiger, Andreas and Leal-Taix{\'e}, Laura and Leibe, Bastian},
    journal=IJCV,
    year=2021
}

@article{dendorfer2021motchallenge,
    title={{MOTChallenge: A Benchmark for Single-Camera Multiple Target Tracking}},
    author={Dendorfer, Patrick and Osep, Aljosa and Milan, Anton and Schindler, Konrad and Cremers, Daniel and Reid, Ian and Roth, Stefan and Leal-Taix{\'e}, Laura},
    journal=IJCV,
    year=2021
}

@article{oquab2023dinov2,
    title={{DINOv2: Learning Robust Visual Features without Supervision}},
    author={Oquab, Maxime and Darcet, Timoth{\'e}e and Moutakanni, Th{\'e}o and Vo, Huy and Szafraniec, Marc and Khalidov, Vasil and Fernandez, Pierre and Haziza, Daniel and Massa, Francisco and El-Nouby, Alaaeldin and others},
    journal=TMLR,
    year=2024
}

@article{simeoni2025dinov3,
    title={{DINOv3}},
    author={Sim{\'e}oni, Oriane and Vo, Huy V. and Seitzer, Maximilian and Baldassarre, Federico and Oquab, Maxime and Jose, Cijo and Khalidov, Vasil and Szafraniec, Marc and Yi, Seungeun and Ramamonjisoa, Micha{\"e}l and Massa, Francisco and Haziza, Daniel and Wehrstedt, Luca and Wang, Jianyuan and Darcet, Timoth{\'e}e and Moutakanni, Th{\'e}o and Sentana, Leonel and Roberts, Claire and Vedaldi, Andrea and Tolan, Jamie and Brandt, John and Couprie, Camille and Mairal, Julien and J{\'e}gou, Herv{\'e} and Labatut, Patrick and Bojanowski, Piotr},
    journal={arXiv preprint arXiv:2508.10104},
    year=2025
}

@inproceedings{dosovitskiy2021vit,
    title={{An Image is Worth 16x16 Words: Transformers for Image Recognition at Scale}},
    author={Alexey Dosovitskiy and Lucas Beyer and Alexander Kolesnikov and Dirk Weissenborn and Xiaohua Zhai and Thomas Unterthiner and Mostafa Dehghani and Matthias Minderer and Georg Heigold and Sylvain Gelly and Jakob Uszkoreit and Neil Houlsby},
    booktitle=ICLR,
    year=2021
}

@inproceedings{chen2023vitadapter,
    title={{Vision Transformer Adapter for Dense Predictions}},
    author={Zhe Chen and Yuchen Duan and Wenhai Wang and Junjun He and Tong Lu and Jifeng Dai and Yu Qiao},
    booktitle=ICLR,
    year=2023
}

@inproceedings{loshchilov2019adamw,
    title={{Decoupled Weight Decay Regularization}},
    author={Loshchilov, Ilya and Hutter, Frank},
    booktitle=ICLR,
    year=2019
}

@inproceedings{dao2024flashattention2,
    title={{FlashAttention-2: Faster Attention with Better Parallelism and Work Partitioning}},
    author={Tri Dao},
    booktitle=ICLR,
    year=2024
}

@inproceedings{ansel2024pytorch2,
    title ={{PyTorch 2: Faster Machine Learning Through Dynamic Python Bytecode Transformation and Graph Compilation}},
    author = {Ansel, Jason and Yang, Edward and He, Horace and Gimelshein, Natalia and Jain, Animesh and Voznesensky, Michael and Bao, Bin and Bell, Peter and Berard, David and Burovski, Evgeni and others},
    booktitle=ASPLOS,
    year=2024
}

@misc{meta2023fvcore,
    title={{fvcore}},
    author={Meta Research},
    year=2023
}

@misc{pytorch_grucell,
    title={{GRUCell --- PyTorch 2.12 Documentation}},
    author={{PyTorch Contributors}},
    howpublished ={\url{https://docs.pytorch.org/docs/2.9/generated/torch.nn.GRUCell.html}},
    note={Accessed: 2026-06-24}
}

@article{zoran2025rvm,
    title={{Recurrent Video Masked Autoencoders}},
    author={Zoran, Daniel and Parthasarathy, Nikhil and Yang, Yi and Hudson, Drew A. and Carreira, Joao and Zisserman, Andrew},
    journal={arXiv preprint arXiv:2512.13684},
    year=2025
}

@article{hochreiter1997long,
    title={{Long Short-Term Memory}},
    author={Hochreiter, Sepp and Schmidhuber, J{\"u}rgen},
    journal={Neural Computation},
    year=1997
}

@inproceedings{smith2023simplified,
    title={{Simplified State Space Layers for Sequence Modeling}},
    author={Smith, Jimmy T. H. and Warrington, Andrew and Linderman, Scott},
    booktitle=ICLR,
    year=2023
}

@inproceedings{cho2014gru,
    title={{Learning Phrase Representations using {RNN} Encoder-Decoder for Statistical Machine Translation}},
    author ={Cho, Kyunghyun and van Merri{\"e}nboer, Bart and Gulcehre, Caglar and Bahdanau, Dzmitry and Bougares, Fethi and Schwenk, Holger and Bengio, Yoshua},
    booktitle=EMNLP,
    year=2014
}

@article{williams1990tbptt,
    title={{An Efficient Gradient-Based Algorithm for Online Training of Recurrent Network Trajectories}},
    author={Williams, Ronald J. and Peng, Jing},
    journal={Neural Computation},
    year=1990
}

@inproceedings{ke2022video,
  title={{Video mask transfiner for high-quality video instance segmentation}},
  author={Ke, Lei and Ding, Henghui and Danelljan, Martin and Tai, Yu-Wing and Tang, Chi-Keung and Yu, Fisher},
  booktitle={ECCV},
  pages={731--747},
  year=2022

}

@inproceedings{liu2025livos,
  title={{Livos: Light video object segmentation with gated linear matching}},
  author={Liu, Qin and Wang, Jianfeng and Yang, Zhengyuan and Li, Linjie and Lin, Kevin and Niethammer, Marc and Wang, Lijuan},
  booktitle={CVPR},
  pages={8668--8678},
  year=2025
 
}
}

\clearpage
\maketitlesupplementary

\normalsize

\renewcommand{\thetable}{\Alph{table}}
\setcounter{table}{0}
\renewcommand{\thefigure}{\Alph{figure}}
\setcounter{figure}{0}

\appendix
\section*{Appendix}
\label{appendix}
\paragraph{Table of contents}
\begin{itemize}
\item \S\ref{sec:supp:impl_details}: Implementation Details
\item \S\ref{sec:supp:add_experiments}: Additional Experiments
\item \S\ref{sec:supp:add_ablations}: Additional Ablations
\item \S\ref{sec:supp:qual_results}: Qualitative Results
\end{itemize}

\section{Implementation Details}
\label{sec:supp:impl_details}
\subsection{Training}
Following PMT~\cite{cavagnero2026pmt}, we leverage frozen DINOv3~\cite{simeoni2025dinov3} pre-trained encoders as the default backbone of \oursShort{}, while also reporting results with DINOv2~\cite{oquab2023dinov2} for comparison. The decoder is pretrained for instance segmentation at image level on the COCO dataset, without temporal supervision, following common practice. In particular, we initialize the decoder from PMT publicly available checkpoints. The GRU-based query propagation module is trained from scratch on the target video dataset. If not stated otherwise, we adopt the common protocol for video segmentation training, as implemented in CAVIS, VidEoMT and PMT~\cite{lee2025cavis,norouzi2026videomt,cavagnero2026pmt}.

\subsection{Hyperparameters}
For all experiments, we follow prior works~\cite{lee2025cavis,cavagnero2026pmt,norouzi2026videomt} with respect to precision, input resolution, and the number of training iterations.
For optimization, we use automatic mixed precision and the AdamW optimizer~\cite{loshchilov2019adamw} with a learning rate of $10^{-4}$, a linear warmup over the first 6{,}000 iterations, and polynomial learning rate decay with a power of 0.9.
Specifically, we train with a batch size of 8 on 8 NVIDIA H100 GPUs. Unless stated otherwise, during training, we process each video using temporal chunks of $F=5$ frames within an overall temporal window of $T=15$ frames, \ie $M=3$ chunks. We train for 160k iterations on YouTube-VIS~\cite{yang2019video} (all versions) and OVIS~\cite{qi2022occluded}, 40k iterations on VIPSeg~\cite{miao2022large}, and 20k iterations on VSPW~\cite{miao2021vspw}. All models use 200 learnable queries with the same dimension as the decoder.

For temporal propagation, we use a GRU~\cite{cho2014gru} cell with shared parameters across all object queries. We implement it using PyTorch's \texttt{GRUCell}~\cite{pytorch_grucell}, with both input and hidden dimensions set to the decoder feature dimension. \\

\PAR{Loss.} To supervise the model, we adopt the same objective as Mask2Former~\cite{cheng2022mask2former}. Specifically, we use the classification cross-entropy loss $\mathcal{L}_{\mathrm{ce}}$ for class predictions, and the binary cross-entropy loss $\mathcal{L}_{\mathrm{bce}}$ and Dice loss $\mathcal{L}_{\mathrm{dice}}$ for mask predictions. The total loss is defined as:
\begin{equation}
\mathcal{L}_{\mathrm{tot}} = \lambda_{\mathrm{bce}} \mathcal{L}_{\mathrm{bce}} + \lambda_{\mathrm{dice}} \mathcal{L}_{\mathrm{dice}} + \lambda_{\mathrm{ce}} \mathcal{L}_{\mathrm{ce}},
\end{equation}
where $\lambda_{\mathrm{bce}} = 5.0$, $\lambda_{\mathrm{dice}} = 5.0$, and $\lambda_{\mathrm{ce}} = 2.0$, strictly following Mask2Former~\cite{cheng2022mask2former}. Deep supervision is enabled by default across all decoder layers.

To ensure temporally consistent supervision, we follow the ground-truth matching strategy introduced in DVIS++~\cite{zhang2025dvis++} and adopted by all competitor approaches. Specifically, each ground-truth object is matched to a query when it first appears, and this assignment is kept fixed in subsequent frames. This encourages the same query to represent the same object throughout the video.

\subsection{Evaluation}
During evaluation, we follow the online video segmentation setting and process each video sequentially, one frame at a time. We report efficiency using two metrics: computational cost, measured in GFLOPs, where one GFLOP corresponds to $10^9$ floating-point operations, and runtime speed, measured in frames per second (FPS).

For FPS measurement, we run 100 warm-up iterations and enable FlashAttention v2~\cite{dao2024flashattention2}, automatic mixed precision, and \texttt{torch.compile} with its default settings~\cite{ansel2024pytorch2}. \texttt{QKV} projections in attention layers are fused to improve latency. GFLOPs are computed using \textit{fvcore}~\cite{meta2023fvcore}. All measurements are conducted on a single NVIDIA H100 GPU with PyTorch 2.9.0 and CUDA 12.8, using a batch size of 1. For each benchmark dataset, FPS and GFLOPs metrics are averaged over all frames in the validation set.

\subsection{Tracking metrics}
\label{sec:supp:tracking_metrics}
In addition to standard video instance segmentation metrics, we report specialized tracking metrics in \cref{tab:tracking_quality_comparison} of the main manuscript to evaluate the temporal consistency and identity preservation of~\oursShort{} across frames.
Specifically, we report the ID F1 score (IDF1)~\cite{ristani2016performance}, which measures identity consistency by computing the F1 score between predicted and ground-truth identities over the entire video. We also report Association Accuracy (AssA)~\cite{luiten2021hota}, which evaluates the quality of temporal associations between matched object instances. MT and ML~\cite{wu2006tracking,dendorfer2021motchallenge} measure track completeness, where MT denotes the number of objects tracked for at least 80\% of their lifespan, and ML denotes the number tracked for less than 20\%. Finally, we report the total number of identity switches~\cite{dendorfer2021motchallenge}, where lower values indicate more stable identity preservation over time.

\begin{table*}[t]
    \centering
    \scriptsize
    \renewcommand{\tabcolsep}{3pt}
    \begin{tabularx}{\linewidth}
    {llcc c YYYYY c YYYYY}
    \toprule
    \multirow{2}[2]{*}{Method} &
    \multirow{2}[2]{*}{Backbone} &
    \multirow{2}[2]{*}{Pre-training} &
    \multirow{2}[2]{*}{Encoder} && 
    \multicolumn{5}{c}{YouTube-VIS 2019 \textit{val}~\cite{yang2019video}} &&
    \multicolumn{5}{c}{YouTube-VIS 2021 \textit{val}~\cite{yang2019video}} \\
    \cmidrule{6-10} \cmidrule{12-16}
    &&&&&
    AP & AP\textsubscript{75} & AR\textsubscript{10} & GFLOPs & FPS &&
    AP & AP\textsubscript{75} & AR\textsubscript{10} & GFLOPs & FPS \\
    \midrule

    DVIS++~\cite{zhang2025dvis++} & ViT-Adapter-L~\cite{chen2023vitadapter} & DINOv2 & \twemoji{fire} &&
    67.7 & 75.3 & 73.7 & 846 & 18 &&
    62.3 & 70.2 & 68.0 & 830 & 17
    \\
    DVIS-DAQ~\cite{zhou2024dvisdaq} & ViT-Adapter-L~\cite{chen2023vitadapter} & DINOv2 & \twemoji{fire} &&
    68.3 & 76.1 & 73.5 & 851 & 10 &&
    62.4 & 70.8 & 68.0 & 836 & 10
    \\
    CAVIS~\cite{lee2025cavis} & ViT-Adapter-L~\cite{chen2023vitadapter} & DINOv2 & \twemoji{fire} &&
    68.9 & 76.2 & 73.6 & 838 & 15 &&
    64.6 & 72.5 & 69.3 & 824 & 15
    \\
    LOMM~\cite{lee2025lomm} & ViT-Adapter-L~\cite{chen2023vitadapter} & DINOv2 & \twemoji{fire} &&
    69.1 & 76.5 & 73.5 & 842 & 12 &&
    65.0 & 72.7 & 69.1 & 842 & 12
    \\
    VidEoMT~\cite{norouzi2026videomt} & ViT-L~\cite{dosovitskiy2021vit} & DINOv2 & \twemoji{fire} &&
    68.6 & 75.6 & 73.9 & 566 & 160 &&
    63.1 & 69.3 & 68.1 & 560 & 160
    \\
    VidEoMT~\cite{norouzi2026videomt} & ViT-L~\cite{dosovitskiy2021vit} & DINOv3 & \twemoji{fire} &&
    68.9 & \underline{77.4} & \textbf{74.8} & 566 & 133 &&
    63.2 & 71.6 & 69.1 & 560 & 133
    \\
    
    \midrule
    CAVIS~\cite{lee2025cavis} & ViT-Adapter-L~\cite{chen2023vitadapter} & DINOv2 & \twemoji{2744} &&
    68.5 & 75.8 & 73.5 & 838 & 15 &&
    64.3 & 72.0 & 68.9 & 824 & 15
    \\
    PMT~\cite{cavagnero2026pmt}  & ViT-L~\cite{dosovitskiy2021vit} & DINOv2 & \twemoji{2744} &&
    68.8 & 75.2 & 73.9 & 617 & 129 &&
    63.8 & 69.4 & 68.1 & 616 & 129
    \\
    \textbf{\oursShort{}} (Ours) & ViT-L~\cite{dosovitskiy2021vit} & DINOv2 & \twemoji{2744} &&
    \textbf{70.1} & \textbf{78.3} & 74.4 & 618 & 127 &&
    \textbf{66.4} & \textbf{75.3} & \textbf{70.4} & 617 & 127
    \\

    \midrule
    CAVIS~\cite{lee2025cavis} & ViT-Adapter-L~\cite{chen2023vitadapter} & DINOv3 & \twemoji{2744} &&
    68.8 & 75.6 & 73.3 & 838 & 13 &&
    63.9 & 71.6 & 68.2 & 824 & 13
    \\
    PMT~\cite{cavagnero2026pmt}  & ViT-L~\cite{dosovitskiy2021vit} & DINOv3 & \twemoji{2744} &&
    69.2 & 76.5 & 74.6 & 617 & 124 &&
    64.3 & 71.2 & 69.0 & 616 & 124
    \\
    \textbf{\oursShort{}} (Ours) & ViT-L~\cite{dosovitskiy2021vit} & DINOv3 & \twemoji{2744} &&
    \underline{69.6} & 76.0 & \underline{74.7} & 618 & 122 &&
    \underline{65.7} & \underline{73.0} & \underline{69.9} & 617 & 122
    \\

    \bottomrule
    \end{tabularx}
    \caption{\textbf{\oursShort{} for VIS on YouTube-VIS 2019 and 2021~\cite{yang2019video}.} }
    \label{tab:sota_comparison_ytvis19_ytvis21}
\end{table*}

\section{Additional Experiments}
\label{sec:supp:add_experiments}
\subsection{Comparison with State-of-the-Art Models}
\label{sec:supp:ytvis1921_comparison}

In the main paper, we compare~\oursShort{} to state-of-the-art methods on the OVIS~\cite{qi2022occluded} and YouTube-VIS 2022~\cite{yang2019video} benchmarks (\cref{tab:sota_comparison_ovis_ytvis22}).
In \cref{tab:sota_comparison_ytvis19_ytvis21}, we further evaluate our method on the YouTube-VIS 2019 and 2021 validation sets.
As in the main paper, we compare against methods that either fine-tune the encoder or keep it frozen.
\oursShort{} achieves high accuracy also on YouTube-VIS 2019 and 2021, although the absolute gains on these datasets are smaller than the ones on more challenging benchmarks. This is expected, as YouTube-VIS 2019 and 2021 are less challenging and already more saturated, leaving less room for improvement.

Compared with finetuned encoder methods that achieve state-of-the-art performance,~\oursShort{} not only achieves superior accuracy but is also substantially more efficient. Equipped with DINOv2 pre-training, it outperforms the most accurate finetuned method, LOMM~\cite{lee2025lomm}, by $+1.0$ AP on YouTube-VIS 2019 and $+1.4$ AP on YouTube-VIS 2021, while being more than $10\times$ faster.

Compared with the most efficient competitor, VidEoMT~\cite{norouzi2026videomt},~\oursShort{} consistently improves accuracy across both DINOv2 and DINOv3 pre-trainings, while maintaining comparable computational cost. A direct comparison with PMT~\cite{cavagnero2026pmt} further isolates the effect of our temporal propagation mechanism. Under the same frozen-encoder setting,~\oursShort{} largely improves over PMT accuracy on both YouTube-VIS 2019 and YouTube-VIS 2021, with almost identical inference speed.

Overall, these results align with the findings in the main paper and show that \oursShort{} improves the accuracy of the strongest efficient baselines while preserving their efficiency across all YouTube-VIS benchmarks.

\begin{table}[t!]
    \centering
    \footnotesize
    \begin{tabularx}{\linewidth}{Xcc}
    \toprule
    GT Metric & High-IDS & Low-IDS \\
    \midrule
    Mean objects per video        & $10$          & $3$         \\
    Disappear rate\textsuperscript{†}(\%) & $63$ & $13$ \\
    Mean disappearance length (frames)   & 13  & 6       \\
    \bottomrule
    \end{tabularx}
    \caption{\textbf{Analysis of high- and low-IDS videos on OVIS \textit{val}.} High- and low-IDS groups are defined as the 20 videos with the highest and lowest number of identity switches, respectively, measured using the GRU-based model from step~\tablestep{1} in~\cref{tab:from_pmt_to_vmt} in the main paper. IDS denotes identity switches. {\textsuperscript{†}Fraction of frames where $\geq$1 object is absent due to occlusion or leaving the scene. All statistics are computed from ground-truth (GT) annotations.}}
    \label{tab:id_switch_analysis}
\end{table}

\subsection{Identity Consistency Analysis}
\label{sec:identity_consistency_analysis}
As discussed in the main paper, adding the GRU to PMT (step~\tablestep{1} in \cref{tab:from_pmt_to_vmt}) substantially improves AP. However, we empirically find that the model still struggles to recover object identities after long occlusions. To better understand this failure mode, we use the predictions of the GRU-based model from step~\tablestep{1} to divide the OVIS validation set into two groups: the 20 videos with the highest number of identity switches and the 20 videos with the lowest number of identity switches. We compute several statistics of these two sets of videos leveraging the corresponding ground-truth annotations, and we report them in \cref{tab:id_switch_analysis}.

The analysis reveals that videos with
crowded scenes and long-term occlusions are considerably more challenging. Compared to the low-IDS group, they contain approximately $3\times$ more objects per video, exhibit nearly $5\times$ higher disappearance rates, and have more than twice the average disappearance length. Notably, the mean disappearance length reaches 13 frames, which is $2.6\times$ longer than the training horizon of the model at step~\tablestep{1} ($T_{\text{train}}=5$).
This suggests that the model struggles to re-identify objects after long occlusions because it has only been trained on much shorter temporal contexts. This motivates the adoption of a longer training strategy based on \tqpshort{}.

This indicates that the recurrent memory is often required to bridge temporal gaps that are substantially longer than those encountered during training, motivating the longer training strategy introduced by \tqpshort{}.

\begin{table}[t!]
    \centering
    \scriptsize
    \begin{tabularx}{\linewidth}{lcc YYYYY}
    \toprule
    Model & GRU & TQP &
    AP & AP\textsubscript{75} & AR\textsubscript{10} & GFLOPs & FPS \\
    \midrule

    VidEoMT~\cite{norouzi2026videomt} & \xmark & \xmark &
     50.1&  53.9& 55.8 & 934 & 104 \\
     \textbf{\oursShort{}} (Ours) & \xmark & \xmark &
     51.1 &  56.0 & 56.7 & 1014 & 97 \\
     \midrule
     
    VidEoMT~\cite{norouzi2026videomt} & \cmark & \xmark &
    51.8 & 56.5 & 57.3 & 935 & 102 \\
    \textbf{\oursShort{}} (Ours) & \cmark & \xmark &
    52.6 & 56.6 & 58.4 &  1015 & 95  \\
     
    \midrule
    VidEoMT~\cite{norouzi2026videomt} & \xmark & \cmark &
    52.1 & 56.7 &  57.5 & 934 & 104  \\
    \textbf{\oursShort{}} (Ours) & \xmark & \cmark &
    54.5 & 59.6 & 59.9 & 1014 & 97  \\

     \midrule
    VidEoMT~\cite{norouzi2026videomt} & \cmark & \cmark &
    52.4 & 56.8 & 57.8 & 935 & 102 \\
    \textbf{\oursShort{}} (Ours) & \cmark & \cmark &
    56.5 & 61.5 & 61.5 & 1015 & 95 \\
    
    \bottomrule
    \end{tabularx}
    \caption{\textbf{Effect of GRU and \tqpshort{}.} Impact of GRU-based propagation and \tqpshort{} training on VidEoMT~\cite{norouzi2026videomt} and \oursShort{} on OVIS \textit{val} with 10-frame training clips.}
    \label{tab:videomt_vs_vmt_ovis}
\end{table}

\begin{table}[th]
    \centering
    \scriptsize
    \begin{tabularx}{\linewidth}{l Y YYYY}

    \toprule

    Method & Size & AP & Params & GFLOPs & FPS \\

    \midrule

    VidEoMT~\cite{norouzi2026videomt} & \multirow{3}{*}{L} &
     51.9 & 316M & 934 & 104
    \\

    PMT~\cite{cavagnero2026pmt} & &
    52.0 & 358M & 1014 & 97
    \\

    \textbf{\oursShort{}} (Ours) & &
    56.7 & 363M & 1015 & 95
    \\

    \midrule

    VidEoMT~\cite{norouzi2026videomt} & \multirow{3}{*}{B} &
     42.7 & 93M & 304 & 178
    \\
    PMT~\cite{cavagnero2026pmt} & &
    42.9 & 116M & 350 & 160
    \\

    \textbf{\oursShort{}} (Ours) & &
    48.1 & 120M & 351 & 155
    \\

    \midrule

    VidEoMT~\cite{norouzi2026videomt} & \multirow{3}{*}{S} &
     31.4 & 24M & 100 & 227
    \\
    PMT~\cite{cavagnero2026pmt} & &
    31.5 & 29M & 110 & 188
    \\

    \textbf{\oursShort{}} (Ours) & &
    39.1 & 30M & 111 & 182
    \\

    \bottomrule
        
    \end{tabularx}
    \caption{\textbf{Impact of model size on OVIS \textit{val}~\cite{qi2022occluded}.} We compare VidEoMT~\cite{norouzi2026videomt}, PMT~\cite{cavagnero2026pmt}, and \oursShort{} across different model sizes.}
    \label{tab:model_size_ovis}
\end{table}

\subsection{Effect of GRU and \tqpshort{} on VidEoMT}
\label{sec:gru_tqp_videomt}

In~\cref{tab:videomt_vs_vmt_ovis}, we study the effect of GRU-based propagation and \tqpshort{} training on VidEoMT~\cite{norouzi2026videomt}. 

For VidEoMT, adding the GRU improves AP by $+1.7$ points, showing that recurrent memory can strengthen the original query-propagation mechanism. Applying \tqpshort{} without the GRU improves AP by $+2.0$ points, indicating that longer-horizon supervision is also beneficial for VidEoMT. When both components are combined, VidEoMT improves over its baseline by $+2.3$ AP.

These results follow a similar trend to~\oursShort{}, but the gains are substantially smaller. This suggests that GRU-based propagation and~\tqpshort{} can improve VidEoMT, while their benefits are better unlocked by the PMT-style architecture used in~\oursShort{}. We hypothesize that fine-tuning the encoder in VidEoMT on relatively limited video segmentation datasets may increase overfitting and limit the effect of long-horizon temporal supervision.

\subsection{Effect of Model Size}
To evaluate how~\oursShort{} scales with backbone size, \cref{tab:model_size_ovis} reports results for ViT-S/B/L backbones and compares against VidEoMT~\cite{norouzi2026videomt} and PMT~\cite{cavagnero2026pmt}. Across all three model sizes, \oursShort{} consistently achieves higher AP than both baselines while maintaining a similar FPS to PMT.

More notably, the advantage over PMT grows as the backbone becomes smaller. This suggests that the proposed memory mechanism is especially helpful when the visual backbone has less capacity. Understanding why smaller backbones benefit more from the proposed temporal memory is an interesting direction for future work.

\begin{table}[t!]
\centering
\scriptsize
\begin{tabular*}{\linewidth}{@{\extracolsep{\fill}}l c c c c c}
\toprule
Method & Backbone & Encoder & AP & AP\textsubscript{75} & AR\textsubscript{10} \\
\midrule

&  & DINOv2~\cite{oquab2023dinov2} & & & \\
\midrule

PMT~\cite{cavagnero2026pmt} & ViT-L~\cite{dosovitskiy2021vit} & \twemoji{2744} & 51.8 & 57.7 & 56.0 \\
\textbf{\oursShort{}} (Ours) & ViT-L~\cite{dosovitskiy2021vit} & \twemoji{2744} & 55.6 & 61.6 & 60.5 \\
PMT~\cite{cavagnero2026pmt} & ViT-L~\cite{dosovitskiy2021vit} & \twemoji{fire} & 53.8 & 56.0 & 58.8 \\
\textbf{\oursShort{}} (Ours) & ViT-L~\cite{dosovitskiy2021vit} & \twemoji{fire} & 56.5 & 61.2 & 61.7 \\

\midrule

&  & DINOv3~\cite{simeoni2025dinov3}& & & \\
\midrule

PMT~\cite{cavagnero2026pmt} & ViT-L~\cite{dosovitskiy2021vit} & \twemoji{2744} & 52.0 & 56.0 & 57.7 \\
\textbf{\oursShort{}} (Ours) & ViT-L~\cite{dosovitskiy2021vit} & \twemoji{2744} & 56.7 & 61.8 & 61.6 \\
PMT~\cite{cavagnero2026pmt} & ViT-L~\cite{dosovitskiy2021vit} & \twemoji{fire} & 52.1 & 56.1 & 57.6 \\
\textbf{\oursShort{}} (Ours) & ViT-L~\cite{dosovitskiy2021vit} & \twemoji{fire} & 56.1 & 60.7 & 61.2 \\

\bottomrule
\end{tabular*}
\caption{\textbf{Effect of encoder fine-tuning on OVIS \textit{val}~\cite{qi2022occluded}.} We compare frozen and fine-tuned encoders for PMT~\cite{cavagnero2026pmt}and~\oursShort{} using DINOv2 and DINOv3 pre-training.}
\label{tab:ablation_finetuning_ovis}
\end{table}

\subsection{Effect of Encoder Fine-Tuning}
\label{sec:effect_of_fine-tuning}
In all experiments reported in the main manuscript, we keep the ViT encoder frozen. In this section, we study whether end-to-end encoder fine-tuning can further improve performance over the frozen-encoder setting. \cref{tab:ablation_finetuning_ovis} reports this comparison for PMT~\cite{cavagnero2026pmt} and~\oursShort{} equipped with both DINOv2~\cite{oquab2023dinov2} and DINOv3~\cite{simeoni2025dinov3} pre-training.

With DINOv2, fine-tuning the encoder improves AP for both PMT and \oursShort{}. With DINOv3, however, the gain becomes negligible or even negative. This suggests that DINOv3 already provides strong frozen representations, and naive end-to-end fine-tuning may perturb these features rather than improve them. This observation aligns with the current design principle of vision foundation models, where a strong frozen encoder can serve as a reusable representation across different downstream tasks without requiring task-specific fine-tuning~\cite{simeoni2025dinov3,cavagnero2026pmt}.

Importantly, \oursShort{} with a frozen DINOv3 encoder remains the strongest setting overall. It outperforms all PMT variants, as well as its fine-tuned counterpart, while maintaining essentially the same computational cost.

\subsection{Analysis of Temporal Gradient Flow}
\label{sec:gradient_flow_analysis}

In \cref{sec:from_pmt_to_vmt} of the main manuscript, we observed that increasing the training-clip length, $T_{\mathrm{train}}$, from 5 to 10 frames decreases AP from 54.2 to 52.3. We hypothesize that this drop results from vanishing gradients.  To verify our hypothesis, we directly analyze the gradient reaching the propagated query states at different frames.

Let $\mathbf{s}_t \in \mathbb{R}^{Q\times C}$ denote the propagated query state entering frame $t$, where $Q$ and $C$ are the number of object queries and the embedding dimensionality of the object queries, respectively. We compute the total training loss, and then differentiate it with respect to the query state at each frame:

\begin{equation}
\mathbf{g}_{t}
=
\frac{\partial \mathcal{L}_{\mathrm{total}}}
     {\partial \mathbf{s}_{t}}.
\end{equation}

Since the gradient $\mathbf{g}_t$ is a $Q \times C$ tensor, we compute its Frobenius norm to obtain a single measure of the overall learning signal reaching frame $t$:
\begin{equation}
\left\lVert \mathbf{g}_{t} \right\rVert_{\mathrm{F}}
=
\sqrt{
\sum_{q=1}^{Q}
\sum_{c=1}^{C}
\left(g_{t}^{q,c}\right)^{2}
}.
\end{equation}
This scalar value allows a direct comparison of gradient magnitudes across the temporal chain. 

To compare gradient propagation over chains of different lengths, we calculate the relative gradient retention:
\begin{equation}
R
=
\frac{G_{\mathrm{first}}}
{G_{\mathrm{last}}}
\times 100\%,
\end{equation}
where $G_{\mathrm{first}}$ and $G_{\mathrm{last}}$ denote the gradient norms at the earliest and latest states of each backward chain.

The results are presented in \Cref{fig:gradient_flow_tqp}. This figure shows that, without \tqpshort{}, only approximately $5\%$ of the gradient signal at the end of the 10-frame chain reaches the earliest propagated state. This sharp decay means that the impact of the supervision signal on the early query states is very weak, leading to only small weight updates and preventing the model from learning effectively on long videos. These results confirm our hypothesis that the model suffers from a vanishing-gradient problem. 

With \tqpshort{}, the relative gradient retention increases to at least $34\%$ within each five-frame chunk. In other words, the shorter backward paths preserve a substantially stronger learning signal, allowing for more meaningful weight updates. Query values are still carried forward across chunk boundaries to preserve information about tracked objects, while detachment prevents gradients from propagating into preceding chunks. Thus, \tqpshort{} maintains temporal continuity while limiting the length of each backward path, thereby mitigating the long-horizon vanishing-gradient problem.

\providecommand{\figwidth}{1\linewidth}
\providecommand{\figheight}{0.65\linewidth}

\definecolor{cObject}{HTML}{3B76C0}
\definecolor{cPeer}{HTML}{F07C21}
\definecolor{cNoTQP}{HTML}{7A5195}
\definecolor{cInk}{HTML}{000000}
\definecolor{cGrid}{HTML}{E2E1DD}

\begin{figure}[t!]
\centering

\begin{tikzpicture}
\begin{axis}[
  width=\figwidth,
  height=\figheight,
  ymode=log,
    ymin=1,
    ymax=60,
   ytick={1,2,5,10,20,50},
   yticklabels={1,2,5,10,20,50},
  xlabel={Frame index},
  ylabel={Gradient norm (log scale)},
  xlabel near ticks,
  ylabel near ticks,
  xmin=-0.7,
  xmax=10.3,
  ymax=60,
  xtick={0,1,2,3,4,5,6,7,8,9},
  yticklabel style={
    color=cInk,
    font=\small
  },
  xticklabel style={
    color=cInk,
    font=\small
  },
  grid=both,
  grid style={
    draw=cGrid,
    line width=0.5pt
  },
  axis lines=box,
  axis line style={
    draw=cInk,
    line cap=butt,
    line width=0.6pt
  },
  xtick pos=bottom,
  ytick pos=left,
  tick align=outside,
  tick style={
    draw=cInk,
    line width=0.6pt
  },
  label style={
    color=cInk,
    font=\small
  },
  legend cell align=left,
  legend style={
    at={(1.0,0.03)},
    anchor=south east,
    draw=none,
    fill=none,
    row sep=0pt,
    font=\scriptsize,
    text=cInk,
    /tikz/every even column/.append style={column sep=4pt}
  },
]

\addplot[
  draw=cNoTQP,
  solid,
  line width=1.20pt,
  mark=*,
  mark size=1.4pt,
  mark options={fill=cNoTQP, draw=cNoTQP}
]
coordinates {
   (0,1.5) (1,2.9) (2,4.3) (3,5.6) (4,7.4) (5,9.8)
  (6,10.3) (7,14.5) (8,18.5) (9,31.6)
};
\addlegendentry{w/o TQP}

\addplot[
  draw=cObject,
  solid,
  line width=1.20pt,
  mark=*,
  mark size=1.4pt,
  mark options={fill=cObject, draw=cObject}
]
coordinates {
  (0,10.8)(1,12.9) (2,18) (3,21) (4,31.5)
};
\addlegendentry{w/ TQP (chunk 1)}

\addplot[
  draw=cPeer,
  solid,
  line width=1.20pt,
  mark=*,
  mark size=1.4pt,
  mark options={fill=cPeer, draw=cPeer}
]
coordinates {
  (5,13.3) (6,17.5) (7,19) (8,22) (9,33.8)
};
\addlegendentry{w/ TQP (chunk 2)}

\node[
  above,
  font=\footnotesize\bfseries,
  text=cNoTQP
] at (axis cs:0.0,1.8) {1.5};

\node[
  right,
  font=\footnotesize\bfseries,
  text=cNoTQP
] at (axis cs:9,26) {31.6};

\node[
  fill=cNoTQP,
  draw=cNoTQP,
  inner sep=1.6pt
] at (axis cs:9,31.6) {};

\node[
  font=\footnotesize\bfseries,
  color=ForestGreen!70!black
] at (axis cs:5.0,7) {5\%};

\node[
  above,
  font=\footnotesize\bfseries,
  text=cObject
] at (axis cs:0.0,13.0) {10.8};

\node[
  above,
  font=\footnotesize\bfseries,
  text=cObject
] at (axis cs:4,33) {31.5};

\node[
  fill=cObject,
  draw=cObject,
  inner sep=1.6pt
] at (axis cs:4,31.5) {};

\node[
  font=\footnotesize\bfseries,
  color=ForestGreen!70!black
] at (axis cs:1.8,26) {34\%};

\node[
  above,
  font=\footnotesize\bfseries,
  text=cPeer
] at (axis cs:5,14.2) {13.3};

\node[
  above=6pt,
  font=\footnotesize\bfseries,
  text=cPeer
] at (axis cs:8.6,32.0) {33.8};

\node[
  fill=cPeer,
  draw=cPeer,
  inner sep=1.6pt
] at (axis cs:9,33.8) {};

\node[
  font=\footnotesize\bfseries,
  color=ForestGreen!70!black
] at (axis cs:7,25) {39\%};

\end{axis}
\end{tikzpicture}

\caption{\textbf{Temporal gradient flow with and without \tqpshort{}.}~
\tqpshort{} shortens the backward path and increases within-chunk gradient retention from approximately 5\% to 37\%.}
\label{fig:gradient_flow_tqp}
\end{figure}
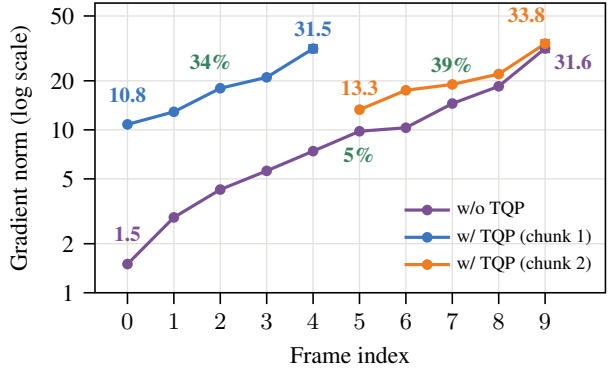

\subsection{GRU Memory Retention Under Occlusion}
\label{sec:gru_memory_retention}
In this analysis, we examine how the GRU's behavior changes when handling occluded objects. Our hypothesis, as mentioned in \cref{sec:gru_prop}, is that the GRU learns to depend more on the queries propagated from the previous frames (\ie, the \textit{memory}), rather than the queries from the current frame, in case an object is occluded in the current frame. We refer to this hypothesized behavior as \textit{memory retention}.

Following the default GRU formulation~\cite{cho2014gru}, the hidden state of the GRU is updated by

\begin{equation}
\mathbf{h}_t
=
\mathbf{z}_t \odot \mathbf{h}_{t-1}
+
(1-\mathbf{z}_t)\odot\mathbf{n}_t,
\label{eq:gru_update}
\end{equation}
where: 
\begin{itemize}
    \item $\mathbf{h}_{t}$ is the GRU's updated \textit{hidden state}, representing the new object queries that will be propagated to the next frame;
    \item $\mathbf{h}_{t-1}$ is the GRU's \textit{hidden state} from the previous time step, which contains the queries propagated from the previous frames;
    \item $\mathbf{n}_t$ is the \textit{candidate state}, which is weighted combination of both the previous hidden state \textit{and} the queries generated in the current frame, controlled by an internal \textit{reset gate};
    \item $\mathbf{z}_t$ is the \textit{update gate}, which determines how much of the previous hidden state is retained and how much is replaced by the candidate state;
    \item $\odot$ denotes the Hadamard product.
\end{itemize}

The update gate $\mathbf{z}_t$ takes values between $0$ and $1$ and controls the balance between the previous hidden state and the candidate state. A value close to $1$ means that the GRU mainly preserves the previous hidden state, whereas a value close to $0$ means that it updates the hidden state using the candidate state. In LVMT, high values for $\mathbf{z}_t$ mean that the new object queries maintain most of the information from the propagated queries, and low values mean that the new object queries are updated with more information from the current frame's queries. This means that we can use $\mathbf{z}_t$ as an indicator of memory retention. If the value for $\mathbf{z}_t$ becomes higher in case of occlusions, this means the model relies more on the memory.

To assess whether this happens, we obtain $\mathbf{z}_t$ for each query and frame, and report its value across different frames in cases with and without occlusions. For this experiment, using the OVIS validation set, we identify occlusion episodes as contiguous spans in which a matched object is fully absent for at least 4 frames. This procedure yields 166 occlusion episodes involving 132 distinct objects across 52 videos.

\providecommand{\figwidth}{\linewidth}
\providecommand{\figheight}{0.78\linewidth}
\definecolor{cObject}{HTML}{3B76C0}
\definecolor{cPeer}{HTML}{2CA25F}
\definecolor{cUnmatched}{HTML}{8A8880}
\definecolor{cInk}{HTML}{000000}
\definecolor{cMuted}{HTML}{8A8880}
\definecolor{cGrid}{HTML}{E2E1DD}
\begin{figure}[t!]
\centering
\begin{tikzpicture}
\begin{axis}[
  width=\figwidth,
  height=\figheight,
  xlabel={Frames from occlusion onset},
  ylabel={Update gate $\mathbf{z}$},
  xlabel near ticks,
  ylabel near ticks,
  xmin=-10,
  xmax=15,
  ymin=0.615,
  ymax=0.780,
  xtick={-10,-5,0,5,10,15},
  ytick={0.63,0.66,0.69,0.72,0.75,0.78},
  yticklabel style={
    color=cInk, font=\small,
    /pgf/number format/fixed,
    /pgf/number format/precision=2
  },
  xticklabel style={color=cInk, font=\small},
  grid=both,
  grid style={draw=cGrid, line width=0.5pt},
  axis lines=box,
  axis line style={draw=cInk, line cap=butt, line width=0.6pt},
  xtick pos=bottom,
  ytick pos=left,
  tick align=outside,
  tick style={draw=cInk, line width=0.6pt},
  label style={color=cInk, font=\small},
  legend cell align=left,
  legend style={
    at={(1.0,0.97)}, anchor=north east,
    draw=none, fill=none, row sep=0pt,
    font=\scriptsize, text=cInk,
    /tikz/every even column/.append style={column sep=4pt}
  },
]
\addplot[draw=none, fill=cMuted, fill opacity=0.20, forget plot]
coordinates {
  (0,0.615) (4,0.615) (4,0.780) (0,0.780)
} \closedcycle;
\draw[draw=cInk, line width=0.6pt] (axis cs:0,0.615) -- (axis cs:0,0.780);
\addplot[draw=cObject, solid, line width=1.20pt, mark=none]
coordinates {
  (-10,0.65224) (-9,0.66487) (-8,0.67881) (-7,0.67123) (-6,0.67270)
  (-5,0.67368) (-4,0.65413) (-3,0.66742) (-2,0.68713) (-1,0.66856)
  (0,0.75151) (1,0.76337) (2,0.76759) (3,0.75681) (4,0.71519)
  (5,0.71125) (6,0.69953) (7,0.70494) (8,0.70124) (9,0.70244)
  (10,0.70688) (11,0.71762) (12,0.71817) (13,0.69505) (14,0.69467)
  (15,0.69238)
};
\addlegendentry{Occluded object}
\addplot[draw=cPeer, densely dashed, line width=0.96pt, mark=none]
coordinates {
  (-10,0.65081) (-9,0.63034) (-8,0.64878) (-7,0.63096) (-6,0.63936)
  (-5,0.64438) (-4,0.65002) (-3,0.64579) (-2,0.65343) (-1,0.63821)
  (0,0.65057) (1,0.65277) (2,0.65396) (3,0.64806) (4,0.64234)
  (5,0.64272) (6,0.64020) (7,0.64186) (8,0.64953) (9,0.64904)
  (10,0.65153) (11,0.64981) (12,0.64889) (13,0.64629) (14,0.64431)
  (15,0.64411)
};
\addlegendentry{Visible objects}
\addplot[draw=cUnmatched, dotted, line width=0.84pt, mark=none]
coordinates {
  (-10,0.69337) (-9,0.69033) (-8,0.70304) (-7,0.69432) (-6,0.69130)
  (-5,0.69353) (-4,0.69910) (-3,0.69963) (-2,0.70454) (-1,0.68979)
  (0,0.71668) (1,0.71955) (2,0.72484) (3,0.72410) (4,0.70410)
  (5,0.70639) (6,0.69828) (7,0.70473) (8,0.70830) (9,0.70526)
  (10,0.71380) (11,0.71141) (12,0.70790) (13,0.69710) (14,0.69962)
  (15,0.70352)
};
\addlegendentry{Unmatched queries}
\node[anchor=north east, font=\scriptsize, text=cInk]
  at (axis cs:-0.4,0.780) {Object disappears};
\node[anchor=south, font=\scriptsize, text=cInk]
  at (axis cs:2,0.615) {Hidden (mean 4 frames)};
\end{axis}
\end{tikzpicture}
\caption{\textbf{GRU memory retention around object occlusion.} Occluded-object queries rely more strongly on memory during the hidden interval.}
\label{fig:gru_z_gate}
\end{figure}
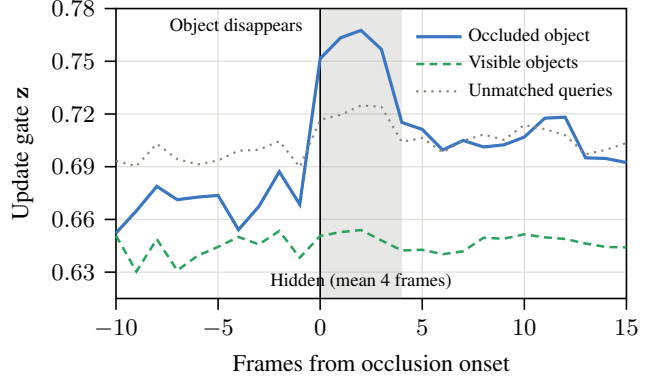

We report the results in \Cref{fig:gru_z_gate}. We find that queries associated with occluded objects retain more of their hidden state when the objects are occluded. Their update-gate value increases from an average of $\sim0.66$ before occlusion to $\sim0.75$ at occlusion onset, corresponding to a relative increase of approximately $10\%$. In comparison, the update-gate values of unmatched queries or queries belonging to visible objects remain virtually unchanged. After the onset response, the update-gate value of the occluded-object queries gradually decreases toward its pre-occlusion level as the objects reappear.

These results show that, when an object becomes occluded, the GRU preserves approximately $10\%$ more information accumulated from previous frames where the object was visible, instead of updating the query from the newly provided frame where the object is occluded. This increased retention allows more stored information to be preserved and propagated to subsequent frames. When the object reappears, the GRU again incorporates more information from the current frame. These results support our hypothesis of \textit{memory retention} under occlusion.

\section{Additional Ablations}
\label{sec:supp:add_ablations}
\subsection{Effect of Chunk Size on \tqpshort}
\cref{tab:ablation_chunk_size_ovis} studies the chunk size $F$ used by \tqpshort{} during training. A chunk size of $F=1$ frame is too small for temporal propagation and gives the weakest result. Larger chunks provide more temporal context and lead to clear gains, with the best performance achieved at $F=5$ frames.

When the chunk size is increased to $F=10$ frames, AP drops to 54.3. This behavior is consistent with what we observe in \cref{tab:ablation_num_frames_gru_TQP_ovis} of the main manuscript, where overly long clips make optimization harder due to vanishing gradients. Therefore, we use $F=5$ frames in all main experiments.

\begin{table}[t!]
    \centering
    \scriptsize
    \begin{tabularx}{\linewidth}{l YYYYY}
    \toprule
    Chunk size &
    AP & AP\textsubscript{75} & AR\textsubscript{10} & GFLOPs & FPS \\
    \midrule

    1 &
    50.9 & 54.2 & 55.4 & 1015 & 95 \\

    3 &
    55.8 & 60.9 & 61.0 & 1015 & 95 \\

    5 &
    56.7 & 61.8 & 61.6 & 1015 & 95 \\

    10 &
    54.3 & 59.7 & 60.0 & 1015 & 95 \\

    \bottomrule
    \end{tabularx}
    \caption{\textbf{Effect of chunk size on OVIS \textit{val}~\cite{qi2022occluded}.} We explore different chunk sizes used by \tqpshort{} during training.}
    \label{tab:ablation_chunk_size_ovis}
\end{table}

\begin{table}[t!]
    \centering
    \scriptsize
    \renewcommand{\tabcolsep}{2.5pt}
    \begin{tabularx}{\linewidth}{lYYY}
    \toprule
    Hidden state initialization &
    AP & AP\textsubscript{75} & AR\textsubscript{10} \\
    \midrule
    Zero Init &
    55.2 & 59.6 & 60.6 \\

    Random Gaussian Init &
    55.8 & 61.5 & 60.8 \\

    Learnable State &
    55.7 & 61.8 & 60.6 \\

    Learnable Object Queries &
    56.7 & 61.8 & 61.6 \\

    \bottomrule
    \end{tabularx}
    \caption{\textbf{Effect of hidden-state initialization on OVIS \textit{val}~\cite{qi2022occluded}.} We evaluate different GRU hidden-state initialization strategies.}
    \label{tab:ablation_hidden_state_init_ovis} 
\end{table}

\subsection{Effect of Hidden-State Initialization.}
\label{sec:ablation_hidden_state_init}
\cref{tab:ablation_hidden_state_init_ovis} compares different strategies to initialize the hidden GRU state. Initializing the hidden state with the shared learnable object queries achieves the best performance.
This result indicates that the hidden state benefits from task-aligned initialization. Since shared object queries are optimized for both object classification and mask prediction, they provide a strong starting point for temporal propagation. In contrast, zero and random initialization lack task-specific information. Having dedicated learnable weights for the hidden state also performs worse, likely because they are less directly coupled to the final prediction objectives.

\section{Limitation}
\paragraph{Training cost.}
TQP bounds peak training memory without affecting inference cost, but its sequential chunk processing increases training time. Specifically, training requires
$M=\lceil T_{\mathrm{train}}/F\rceil$ forward and backward passes per iteration, one for each chunk. Consequently, wall-clock training time and computation increase with the number of chunks. Therefore, developing a method that retains TQP's benefits without additional training time, while preserving inference efficiency, is a valuable direction for future work.

\newcommand{\qualovisframeheight}{4cm}
\newcommand{\qualovisrowlabel}[1]{\parbox[c][\qualovisframeheight][c]{\linewidth}{\centering \scriptsize $t=#1$}}

\begin{figure*}[p]
\centering

\begin{minipage}[t]{0.06\linewidth}
    \centering
    \vspace*{-\qualovisframeheight}
    \qualovisrowlabel{0}
    \qualovisrowlabel{4}
    \qualovisrowlabel{8}
    \qualovisrowlabel{10}
    \qualovisrowlabel{11}
    \caption*{}
\end{minipage}
\begin{minipage}[t]{0.30\linewidth}
    \centering
    \includegraphics[width=\linewidth,height=\qualovisframeheight]{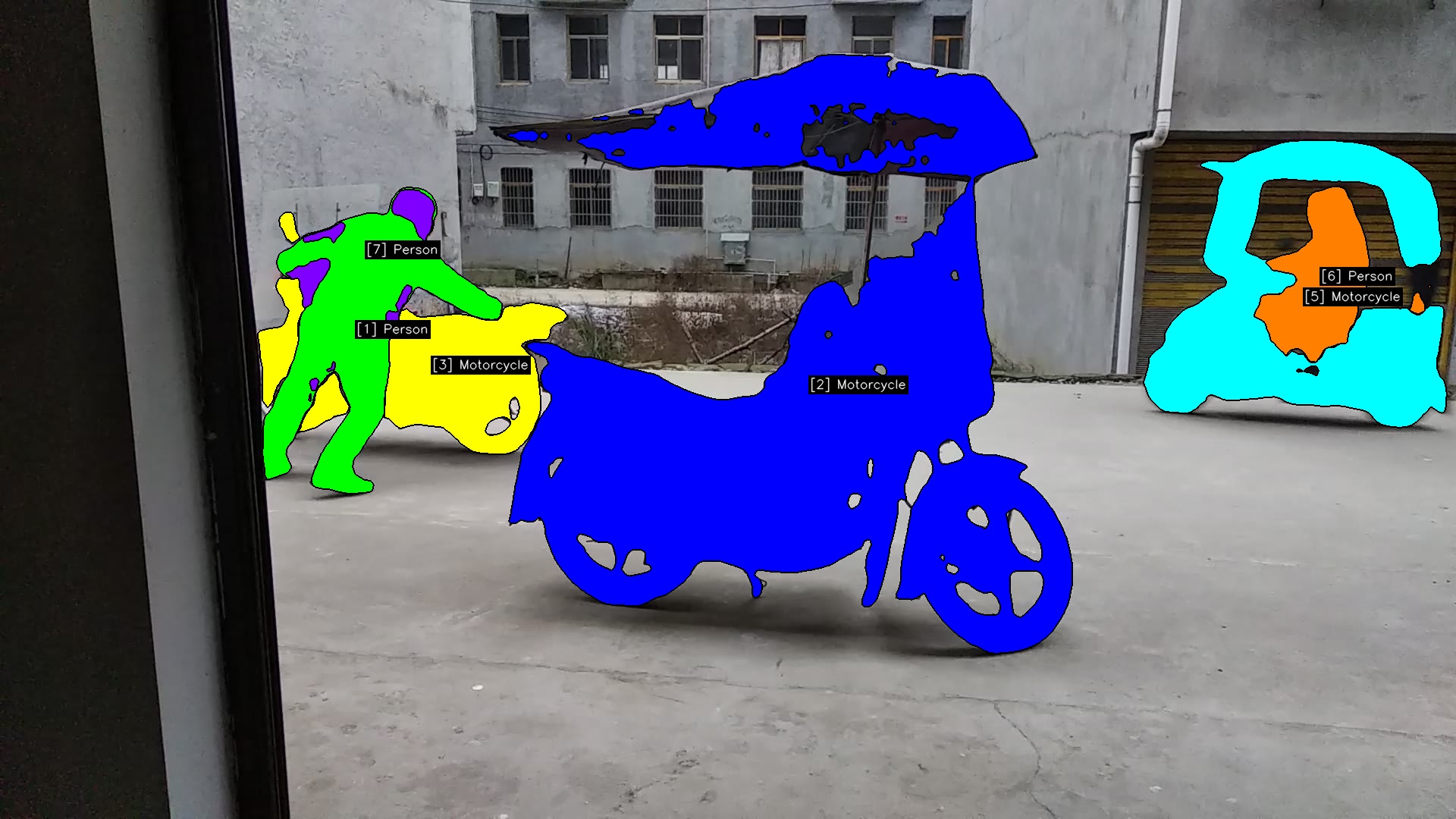}
    \includegraphics[width=\linewidth,height=\qualovisframeheight]{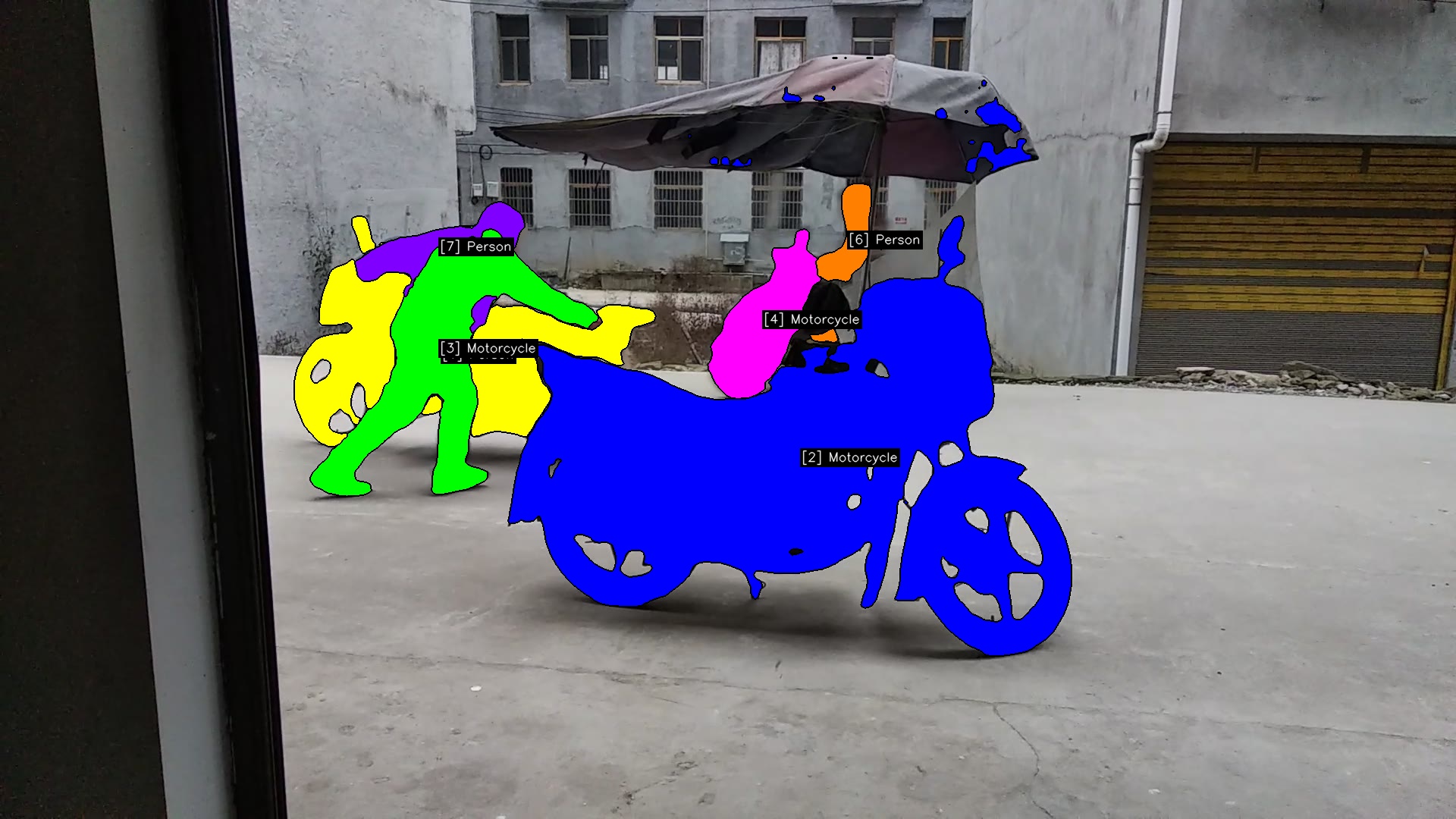}
    \includegraphics[width=\linewidth,height=\qualovisframeheight]{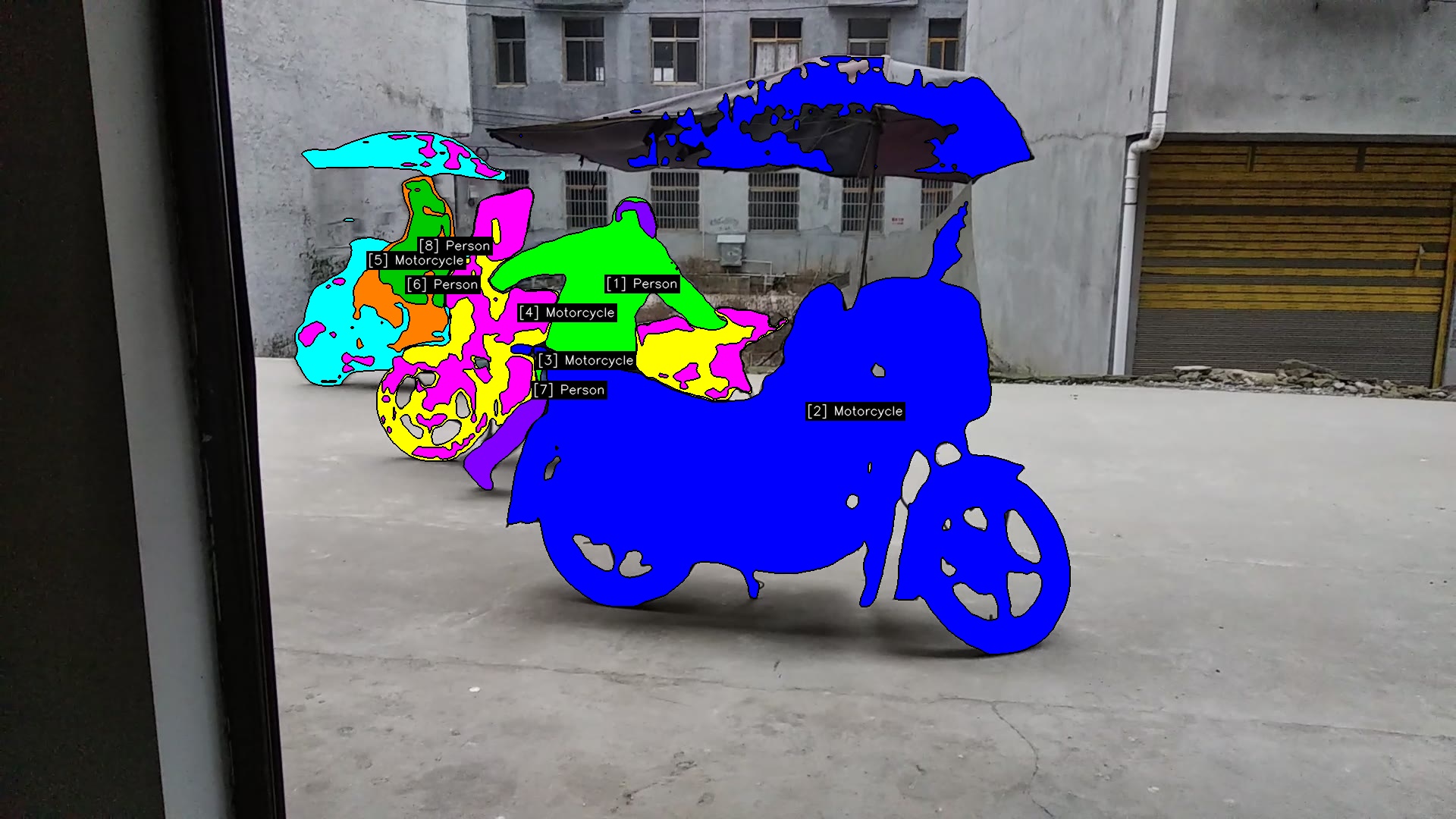}
    \includegraphics[width=\linewidth,height=\qualovisframeheight]{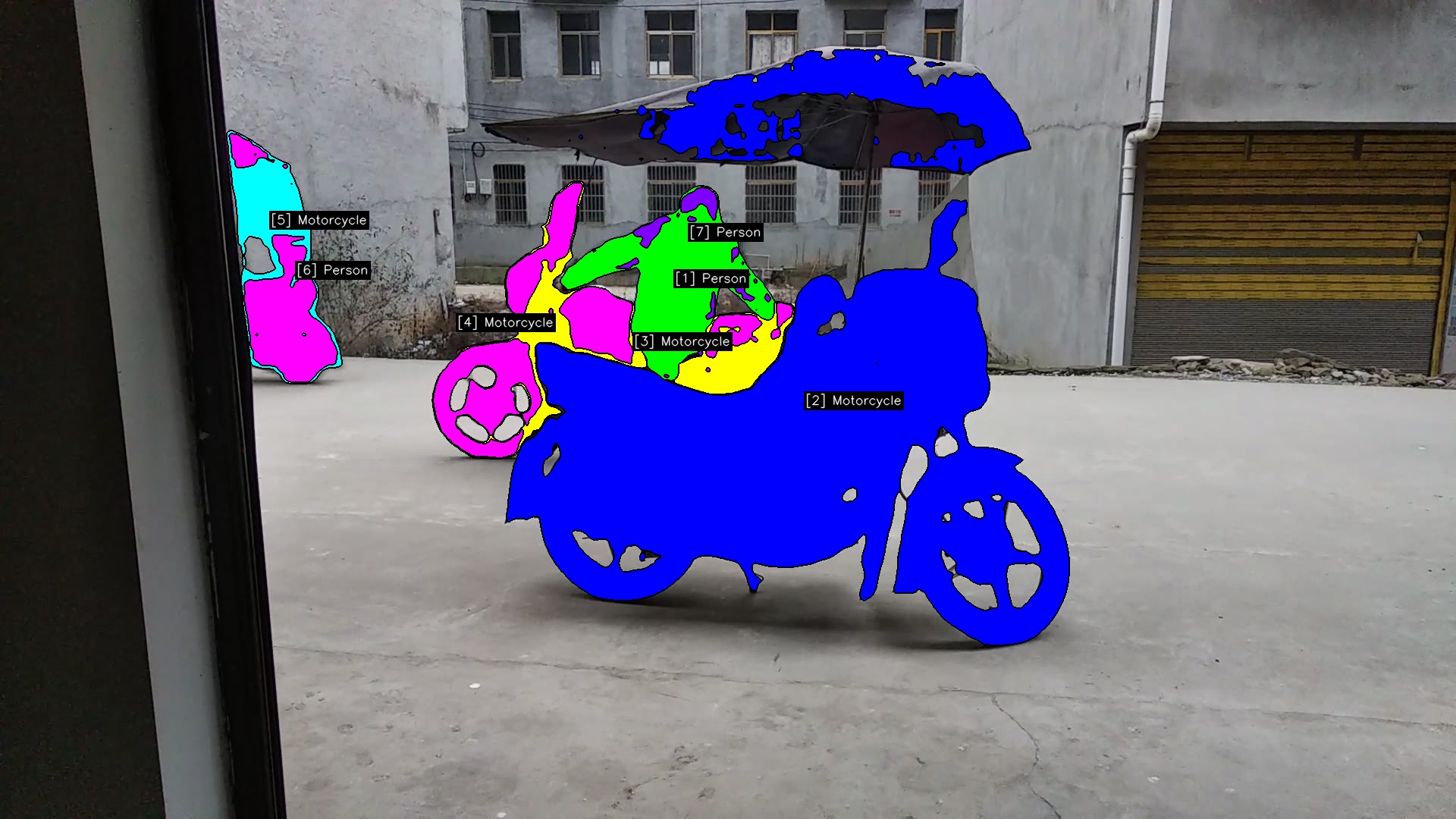}
    \includegraphics[width=\linewidth,height=\qualovisframeheight]{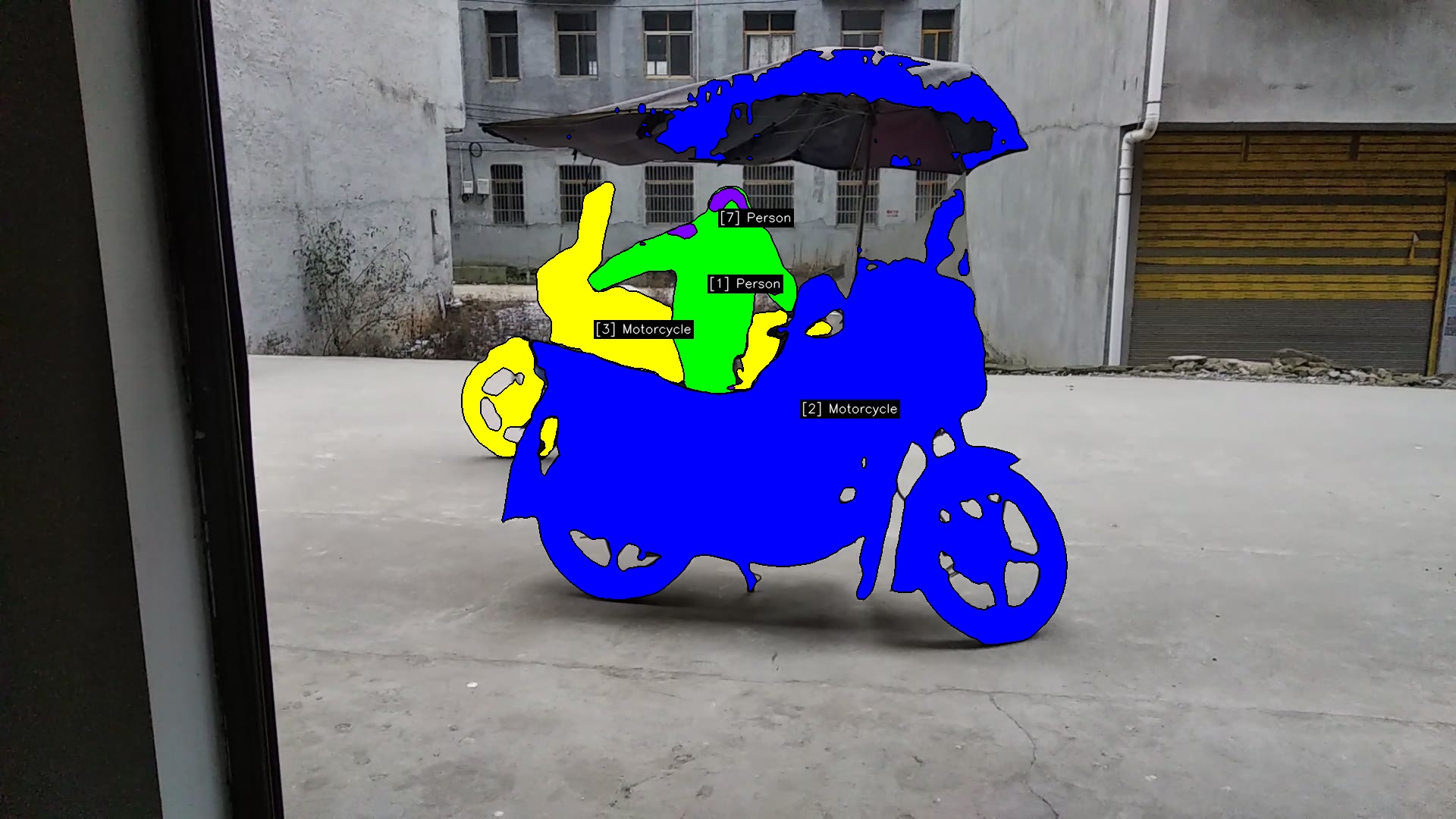}
    \caption*{PMT~\cite{cavagnero2026pmt}}
\end{minipage}
\begin{minipage}[t]{0.30\linewidth}
    \centering
    \includegraphics[width=\linewidth,height=\qualovisframeheight]{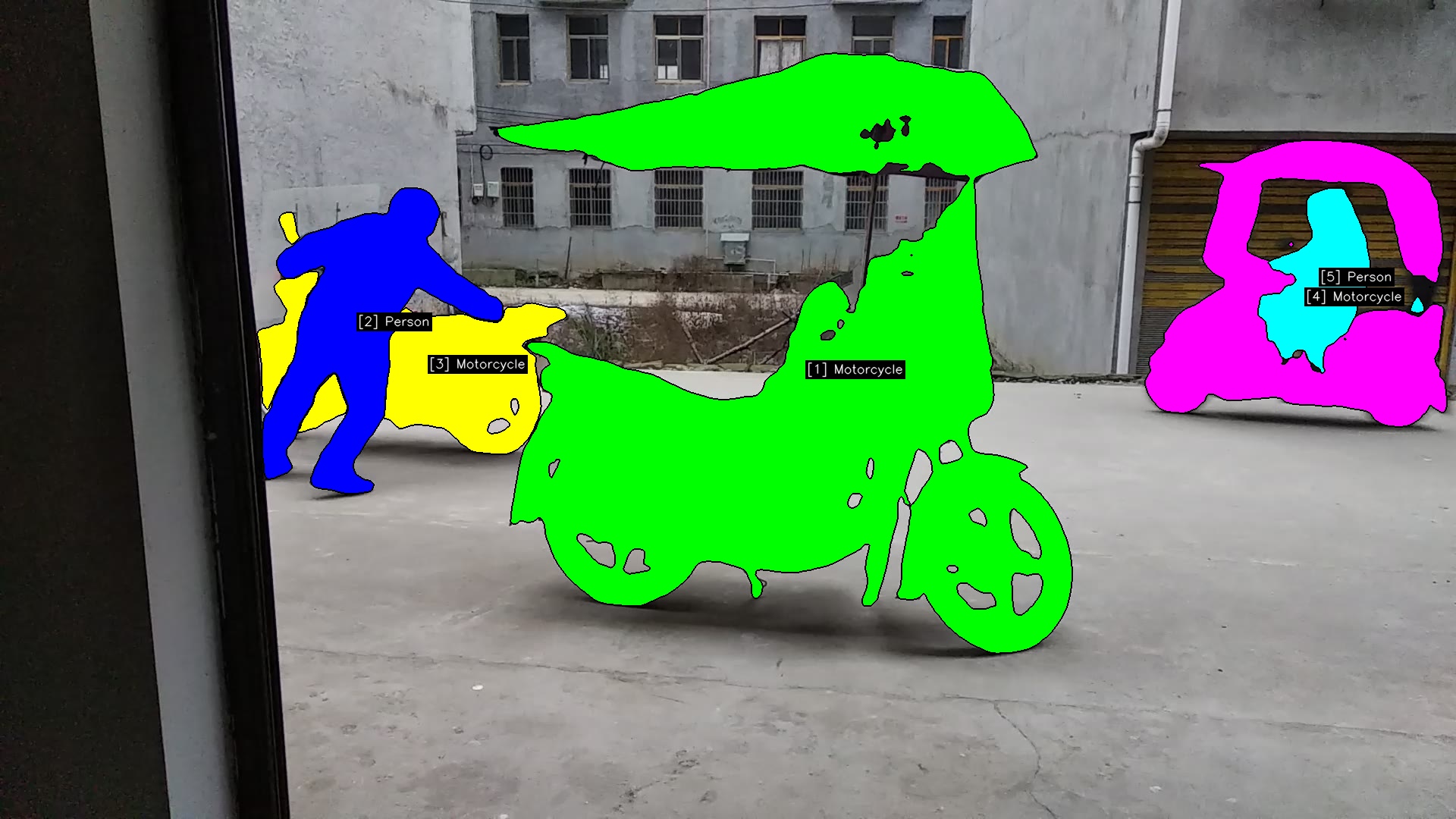}
    \includegraphics[width=\linewidth,height=\qualovisframeheight]{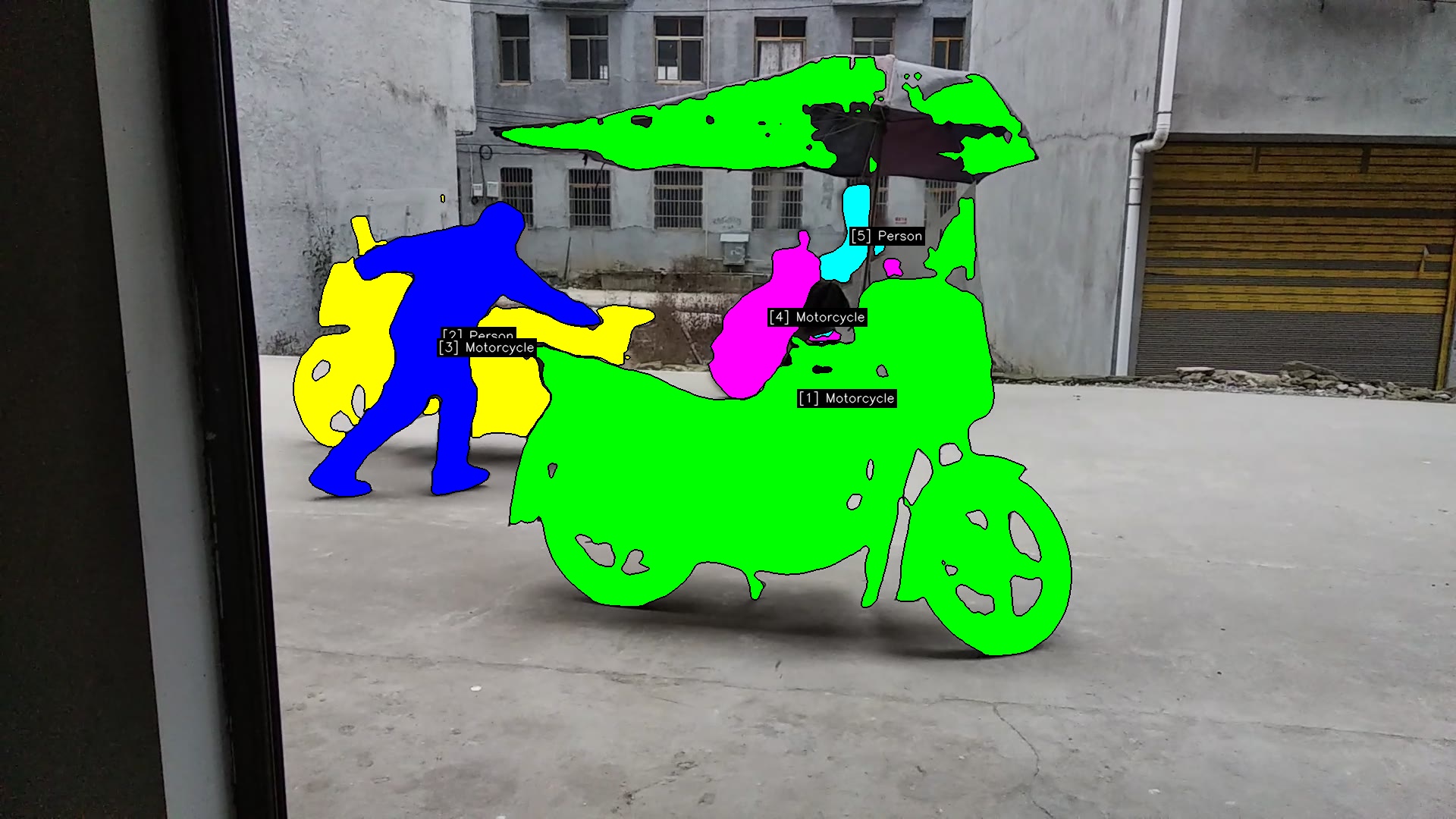}
    \includegraphics[width=\linewidth,height=\qualovisframeheight]{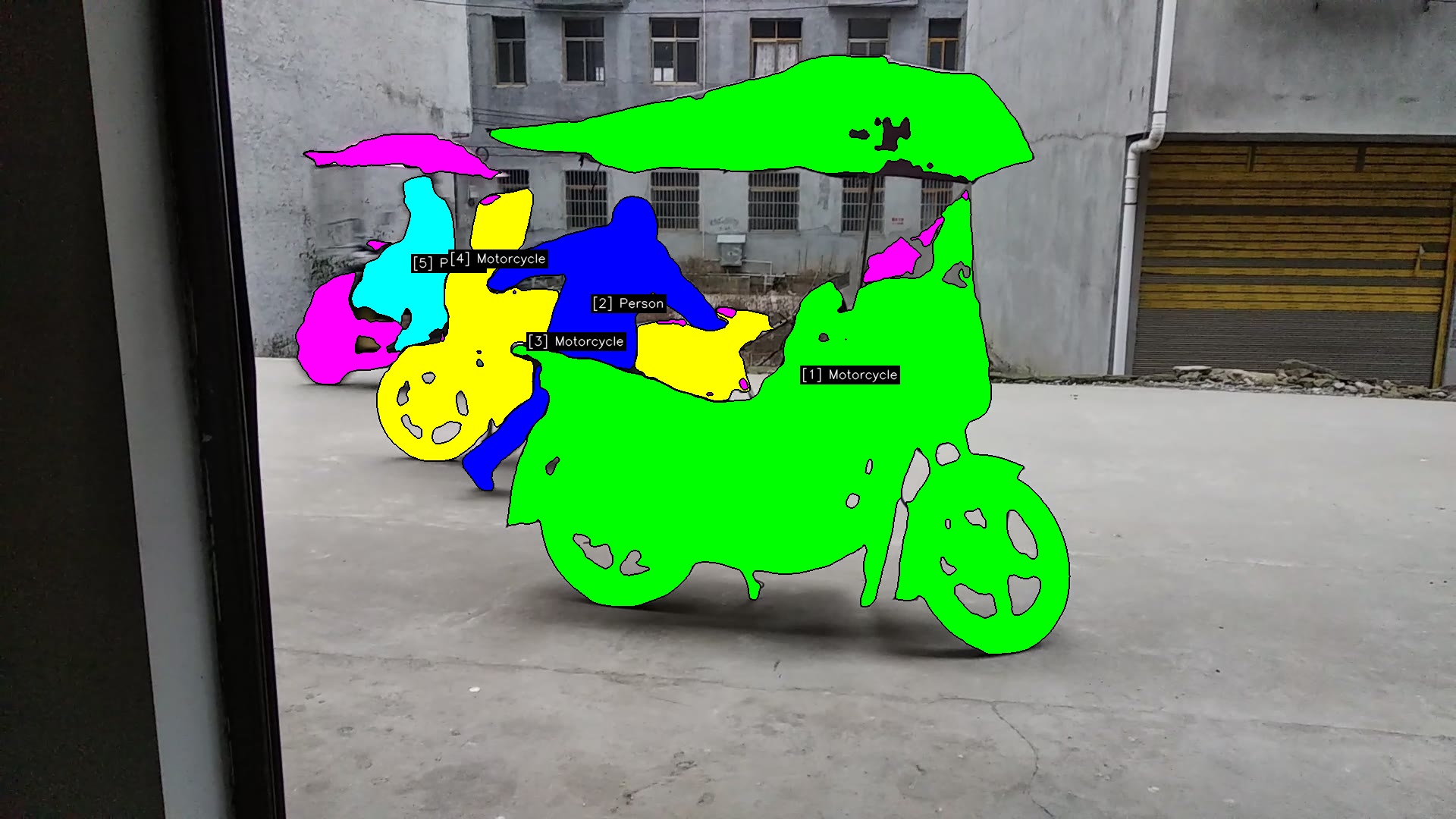}
    \includegraphics[width=\linewidth,height=\qualovisframeheight]{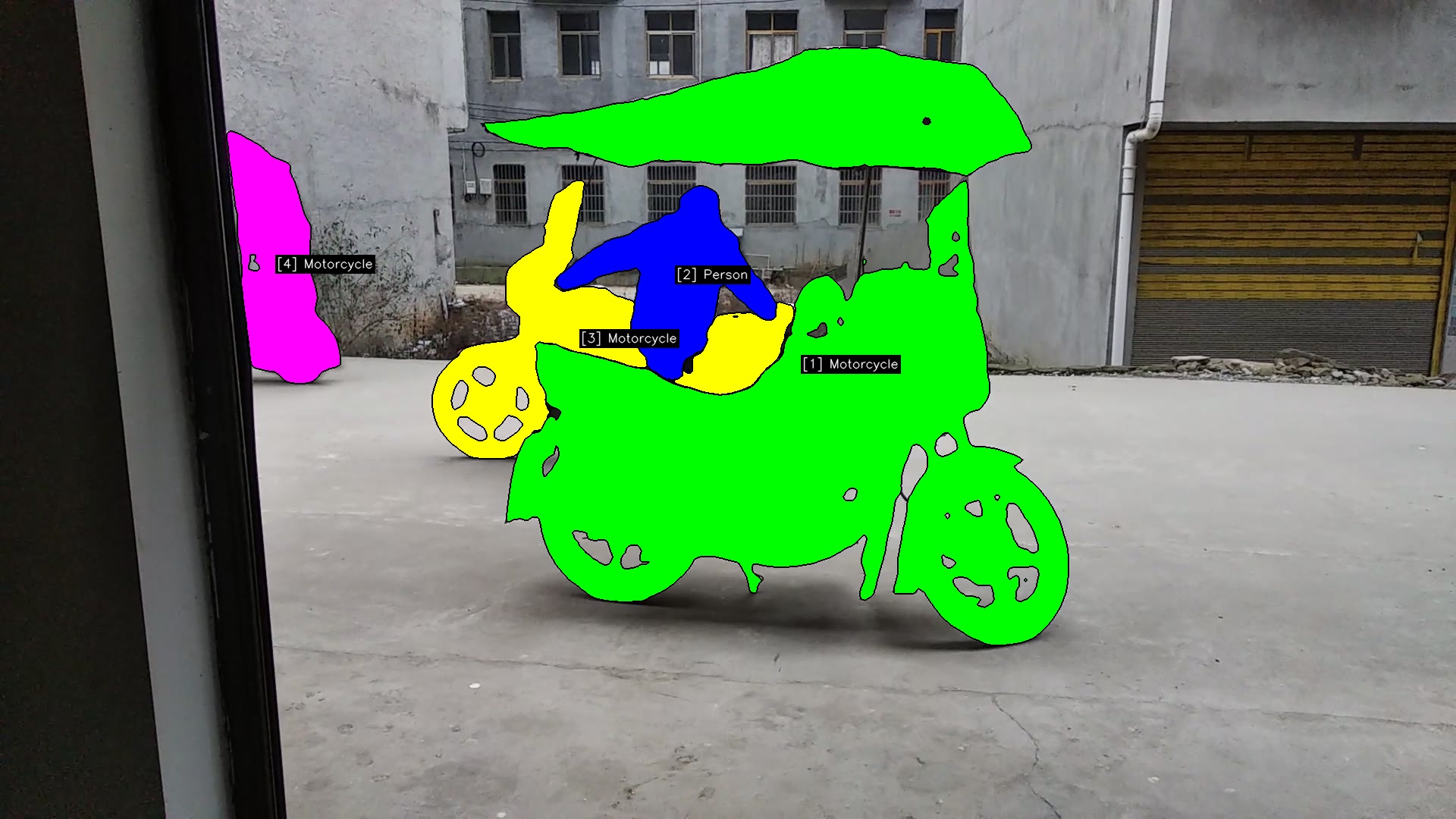}
    \includegraphics[width=\linewidth,height=\qualovisframeheight]{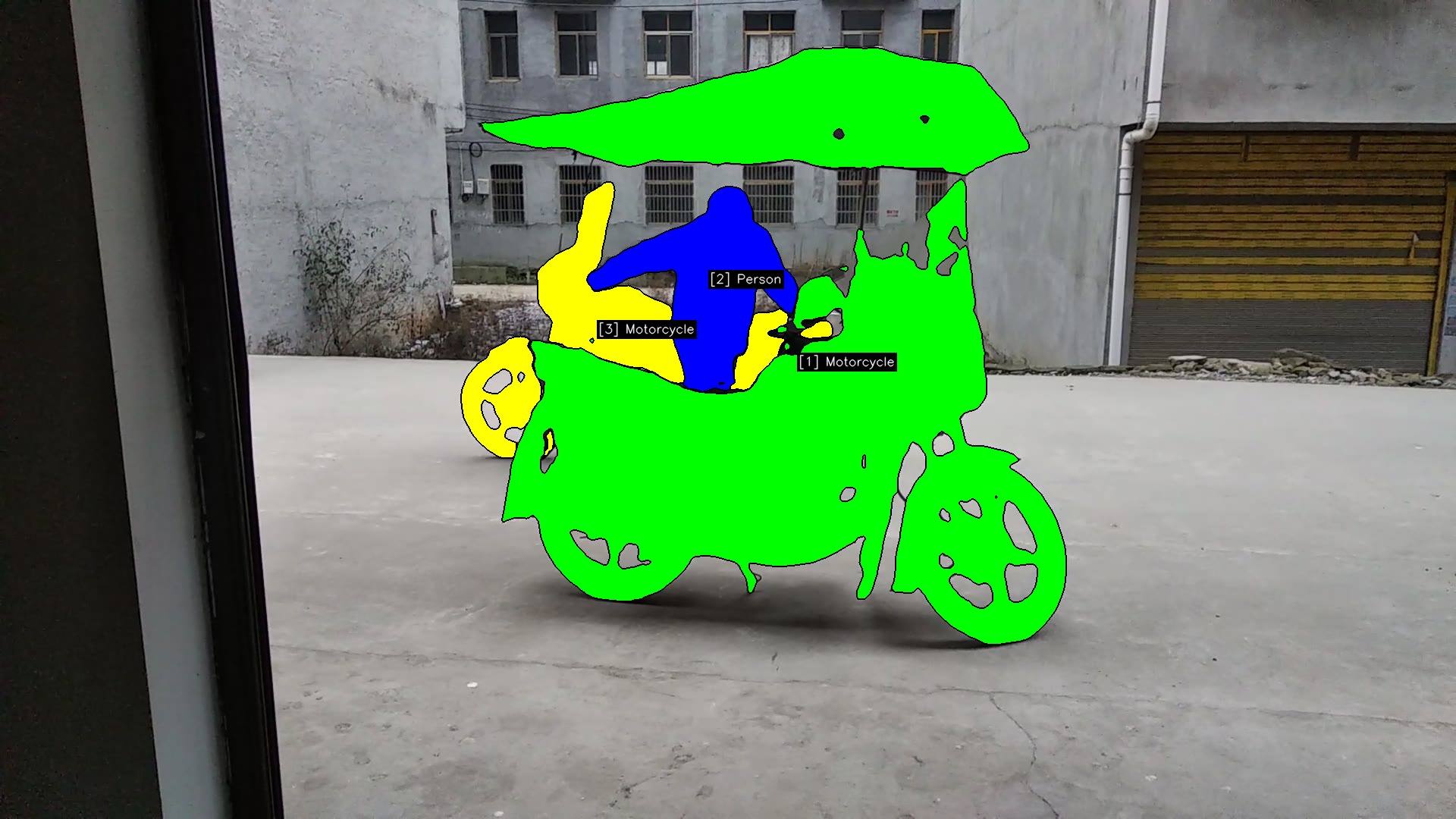}
    \caption*{\oursShort{} (Ours)}
\end{minipage}
\begin{minipage}[t]{0.30\linewidth}
    \centering
    \includegraphics[width=\linewidth,height=\qualovisframeheight]{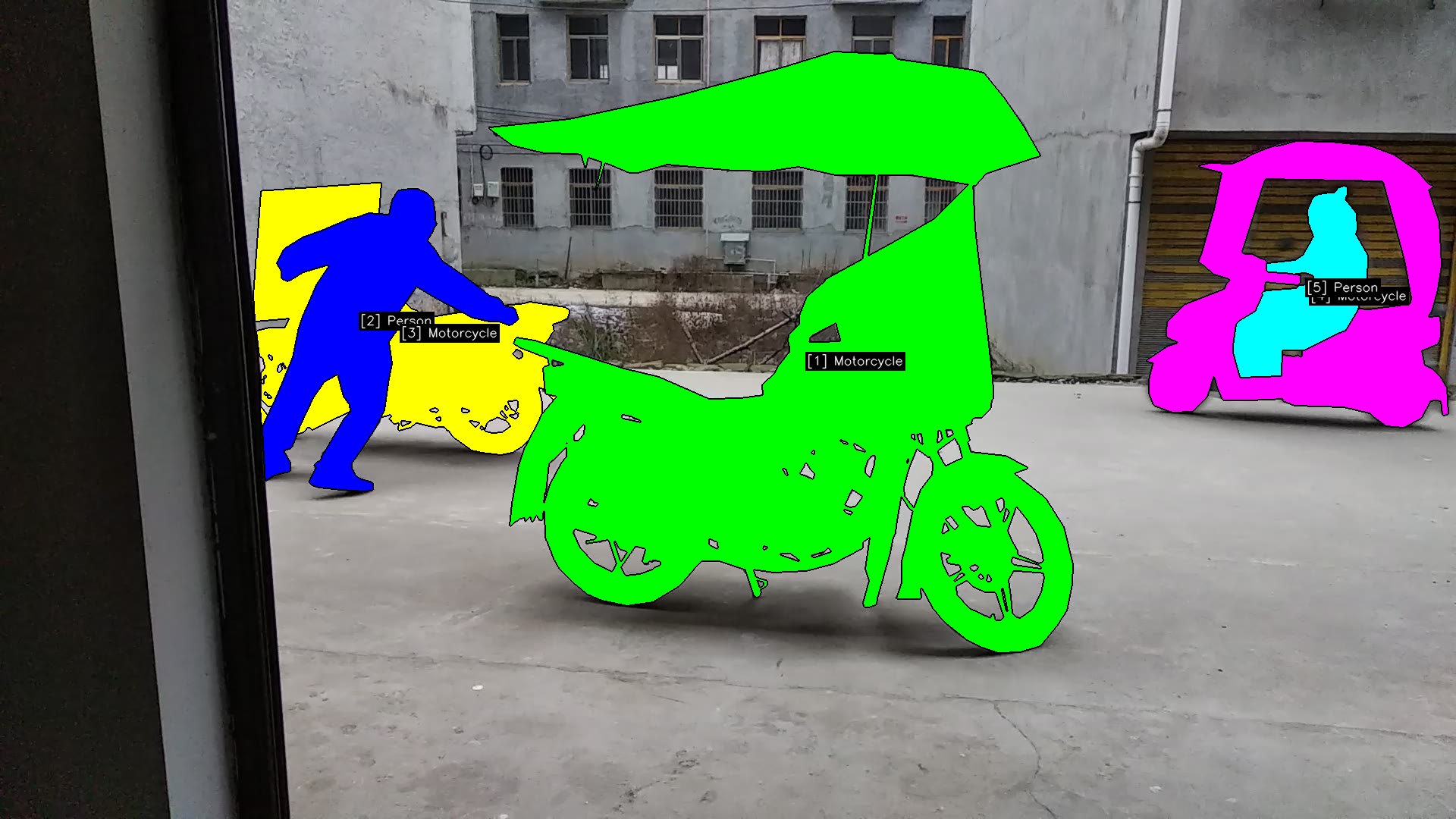}
    \includegraphics[width=\linewidth,height=\qualovisframeheight]{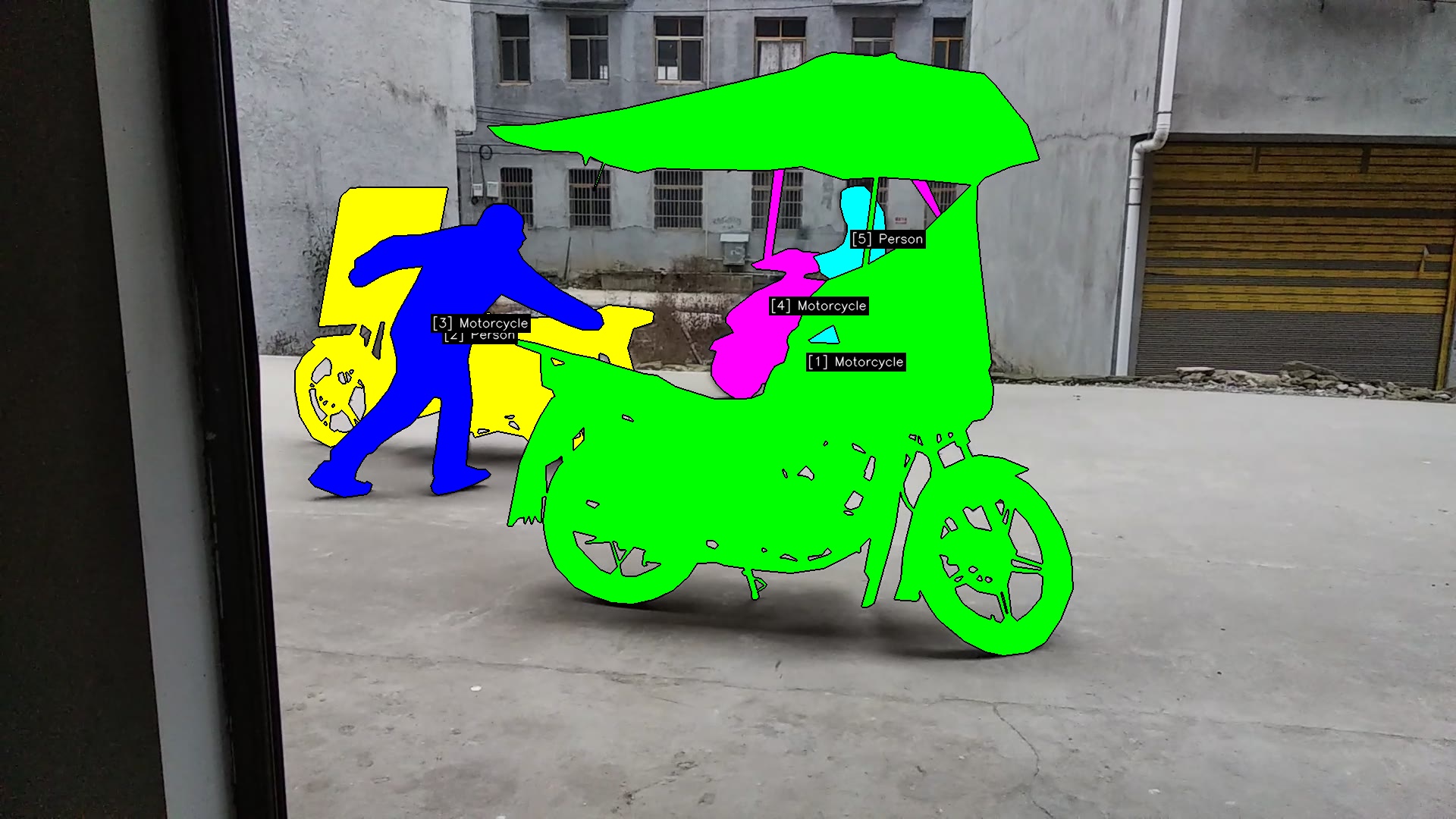}
    \includegraphics[width=\linewidth,height=\qualovisframeheight]{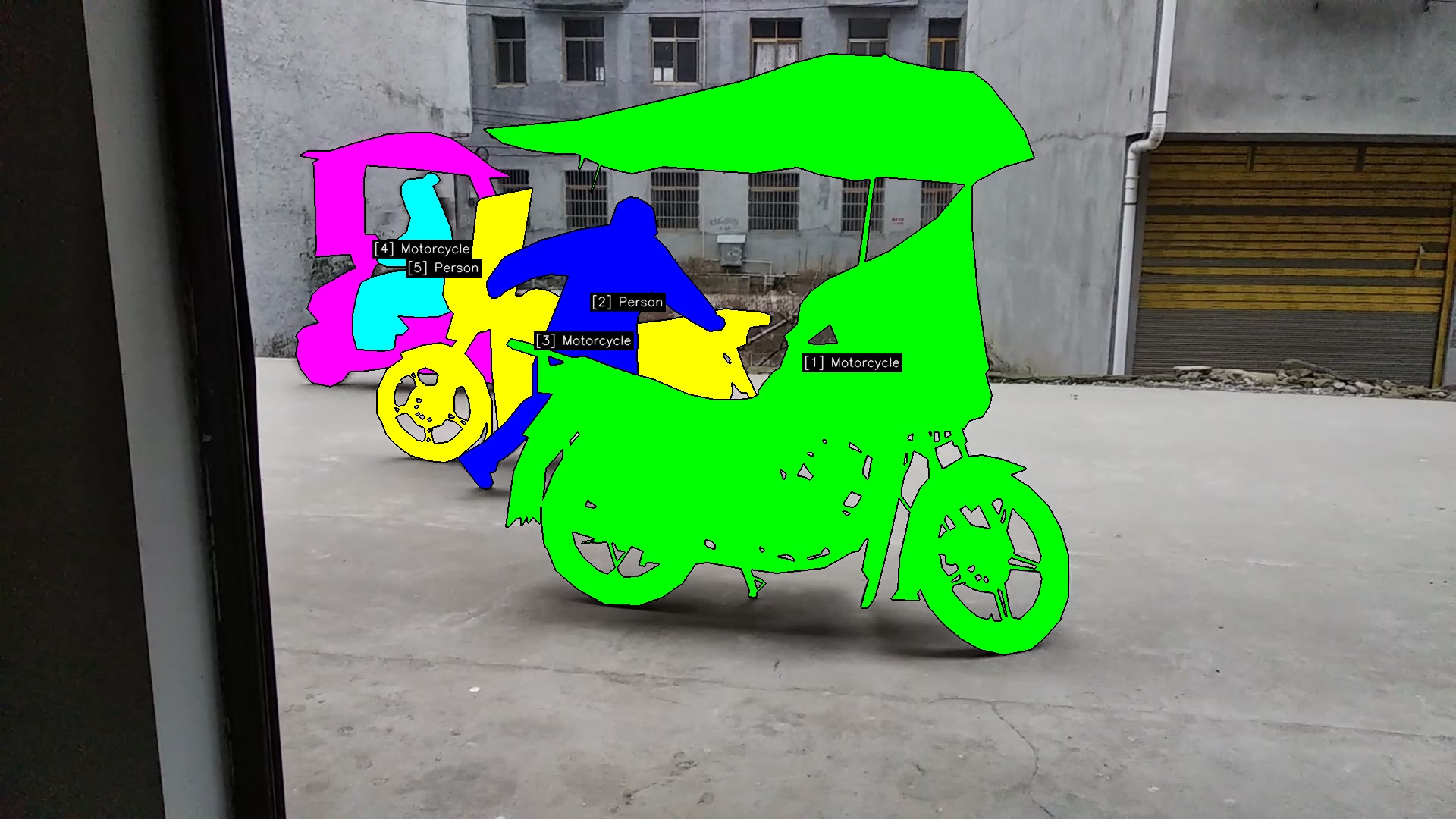}
    \includegraphics[width=\linewidth,height=\qualovisframeheight]{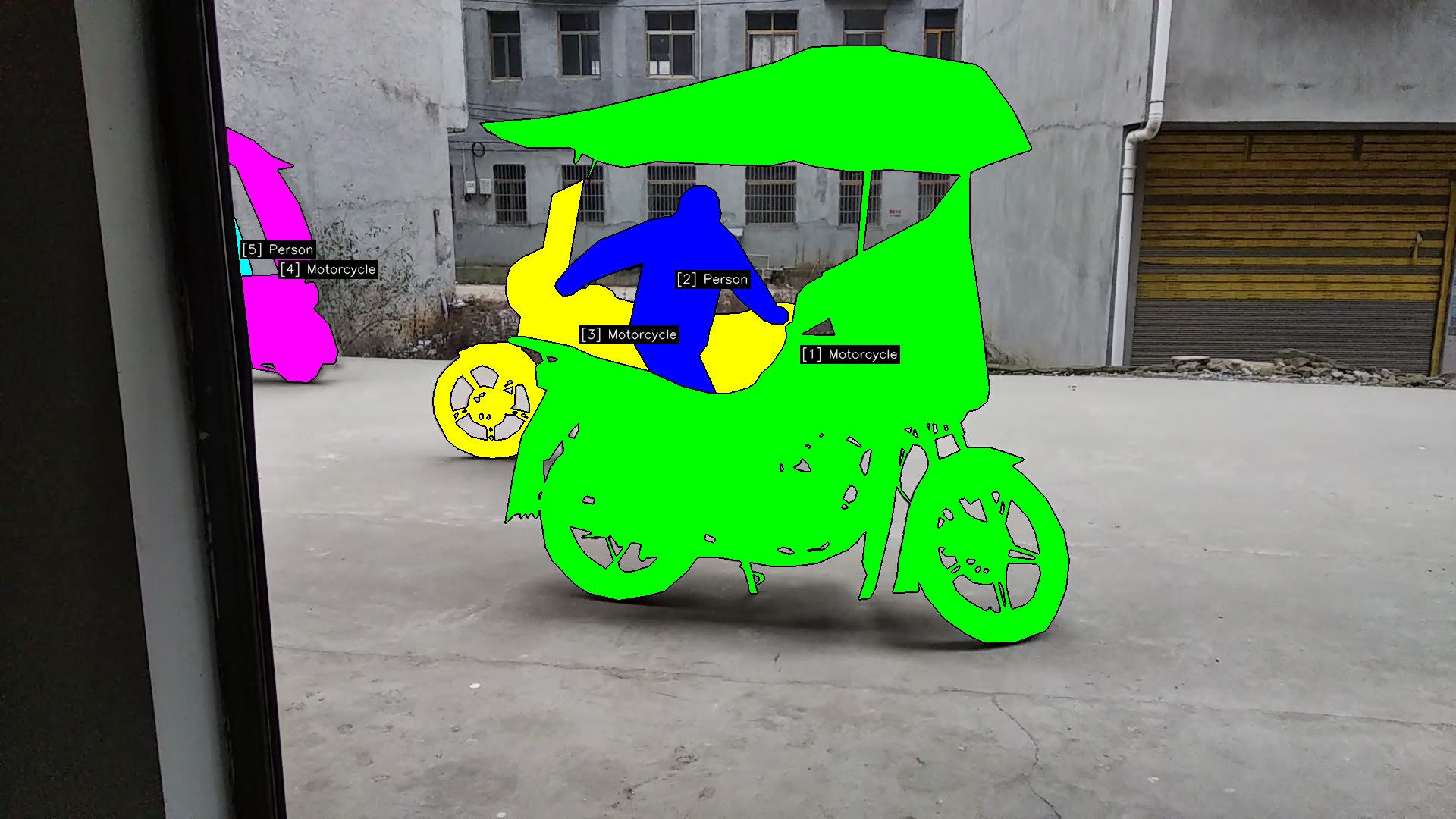}
    \includegraphics[width=\linewidth,height=\qualovisframeheight]{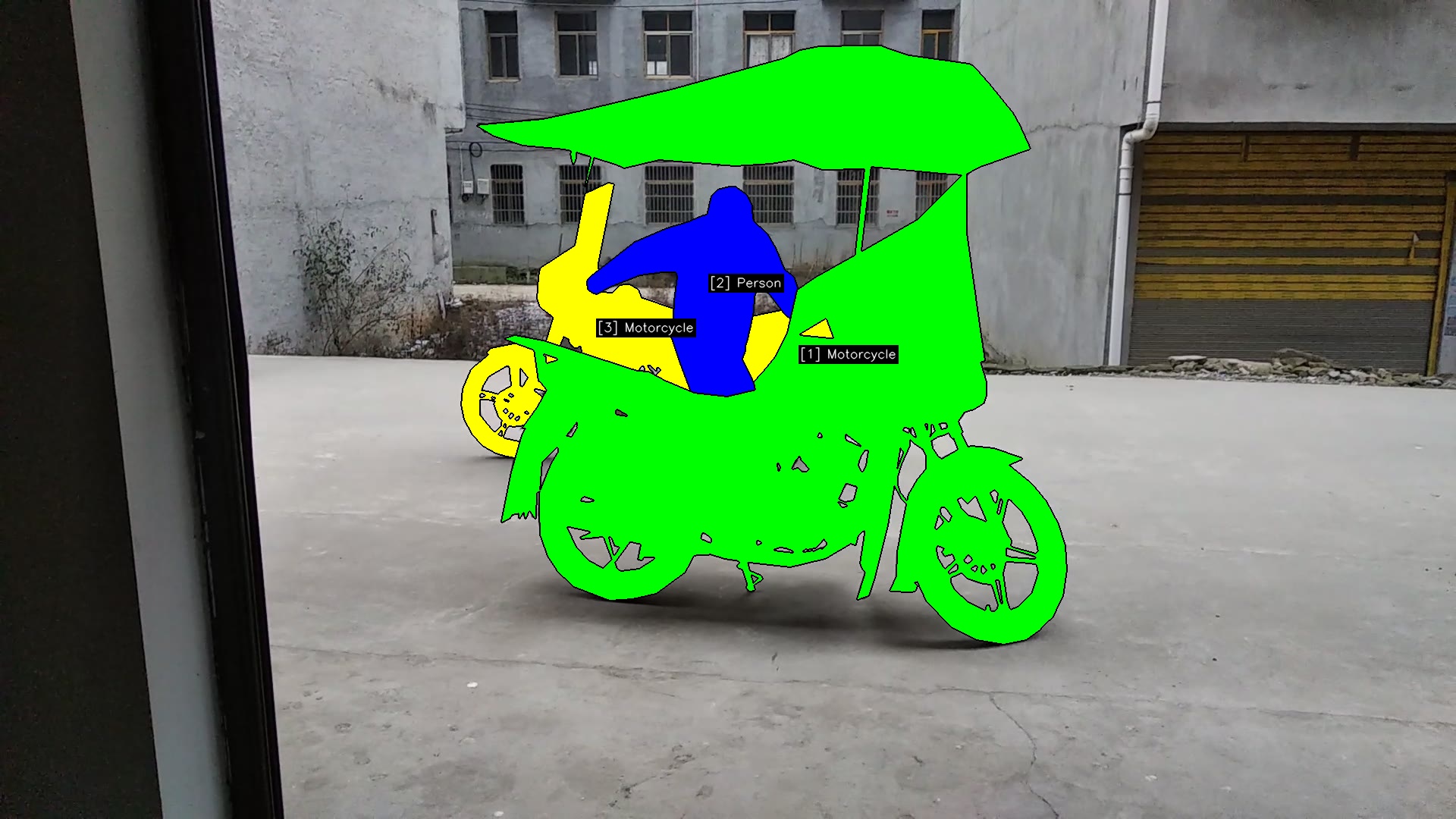}
    \caption*{Ground Truth}
\end{minipage}

\caption{\textbf{Qualitative results on OVIS~\cite{qi2022occluded}.} Comparison between PMT~\cite{cavagnero2026pmt}, \oursShort{}, and the ground-truth annotations on selected frames $t=\{0,4,8,10,11\}$. PMT suffers from identity switches at $t=4$ and $t=8$ under heavy occlusion among similar motorcycles and riders, while \oursShort{} preserves more consistent identities across the sequence.}
\label{app:fig:qual_ovis}
\end{figure*}

\section{Qualitative Results}
\label{sec:supp:qual_results}
\cref{app:fig:qual_ovis} shows a challenging OVIS~\cite{qi2022occluded} video with multiple visually similar motorcycles and riders undergoing heavy occlusion. PMT~\cite{cavagnero2026pmt} starts producing identity switches at frame~4, when one motorcycle becomes occluded, and the errors become more severe at frame~8 under stronger occlusion. In contrast, \oursShort{} preserves the correct identities throughout the sequence.

\end{document}